\documentclass[10pt,twocolumn,letterpaper]{article}

\usepackage[pagenumbers]{cvpr}
\usepackage{graphicx}
\usepackage{amsmath}
\usepackage{amssymb}
\usepackage{booktabs}
\usepackage{enumitem}
\usepackage{caption}
\usepackage{bbm}
\usepackage{multirow}
\usepackage{makecell}
\usepackage{color}
\usepackage{balance}
\usepackage[pagebackref,breaklinks,colorlinks]{hyperref}
\usepackage[capitalize]{cleveref}
\usepackage{xcolor}
\newcommand{\improve}[1]{{\color{blue!20!black!50!green}{\small($\uparrow${#1})}}}
\newcommand{\reduce}[1]{{\color{blue!20!black!30!red}{\small($\downarrow${#1})}}}
\newcommand{\improveb}[1]{{\color{blue!20!black!50!green}{($\downarrow${#1})}}}
\newcommand{\reduceb}[1]{{\color{blue!20!black!30!red}({$\uparrow${#1}})}}
\newcommand{\bm}[1]{\mathbf{#1}}
\setlist[itemize]{noitemsep,leftmargin=*,topsep=0in}
\setlist[enumerate]{noitemsep,leftmargin=*,topsep=0in}
\begin{document}
\title{Learning Articulated Shape with Keypoint Pseudo-labels from Web Images}
\author{Anastasis Stathopoulos\\
Rutgers University\\
\and
Georgios Pavlakos\\
UC Berkeley\\
\and
Ligong Han\\
Rutgers University\\
\and
Dimitris Metaxas\\
Rutgers University\\
}
\maketitle

\begin{abstract}
This paper shows that it is possible to learn models for monocular 3D reconstruction of articulated objects (\eg, horses, cows, sheep), using as few as 50-150 images labeled with 2D keypoints. Our proposed approach involves training category-specific keypoint estimators, generating 2D keypoint pseudo-labels on unlabeled web images, and using both the labeled and self-labeled sets to train 3D reconstruction models. It is based on two key insights: (1) 2D keypoint estimation networks trained on as few as 50-150 images of a given object category generalize well and generate reliable pseudo-labels; (2) a data selection mechanism can automatically create a ``curated'' subset of the unlabeled web images that can be used for training -- we evaluate four data selection methods. Coupling these two insights enables us to train models that effectively utilize web images, resulting in improved 3D reconstruction performance for several articulated object categories beyond the fully-supervised baseline. Our approach can quickly bootstrap a model and requires only a few images labeled with 2D keypoints. This requirement can be easily satisfied for any new object category. To showcase the practicality of our approach for predicting the 3D shape of arbitrary object categories, we annotate 2D keypoints on giraffe and bear images from COCO -- the annotation process takes less than 1 minute per image.
\end{abstract}
\begin{figure}[t]
  \centering
  \includegraphics[width=\columnwidth]{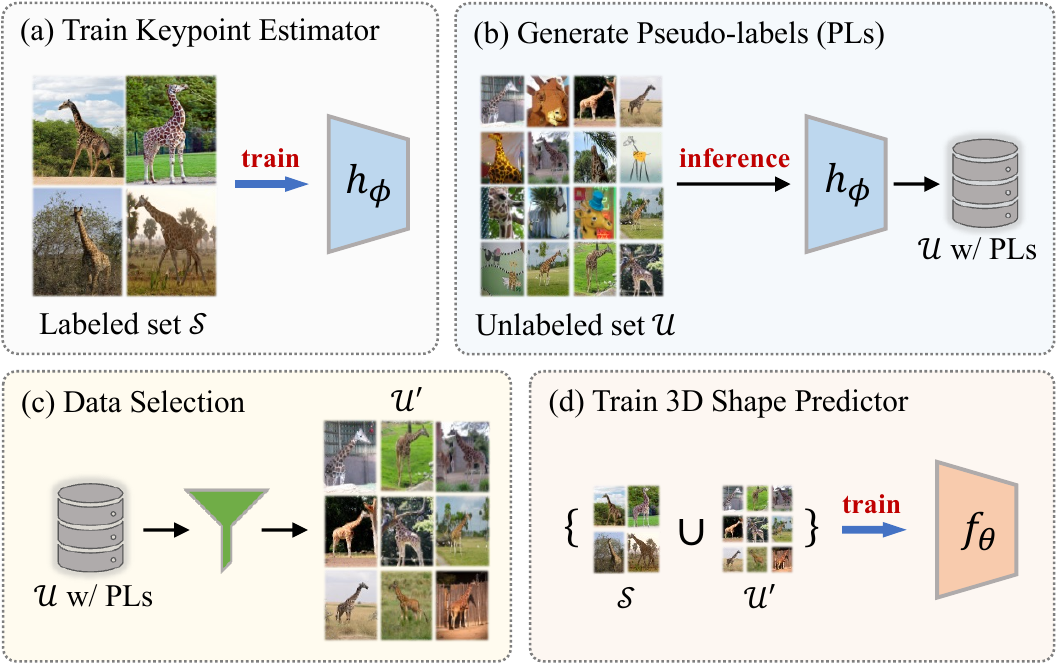}
  \caption{
    {\bf Overview of the proposed framework}.
    It includes: (a) training a category-specific keypoint estimator with a limited labeled set $\mathcal{S}$, (b) generating keypoints pseudo-labels on web images, (c) automatic curation of web images to create a subset $\mathcal{U}'$, and (d) training a model for 3D shape prediction with images from $\mathcal{S}$ and $\mathcal{U}'$.
  }
  \label{fig:intro_fig}
\end{figure}
\begin{figure*}[t]
  \centering
  \includegraphics[width=\textwidth]{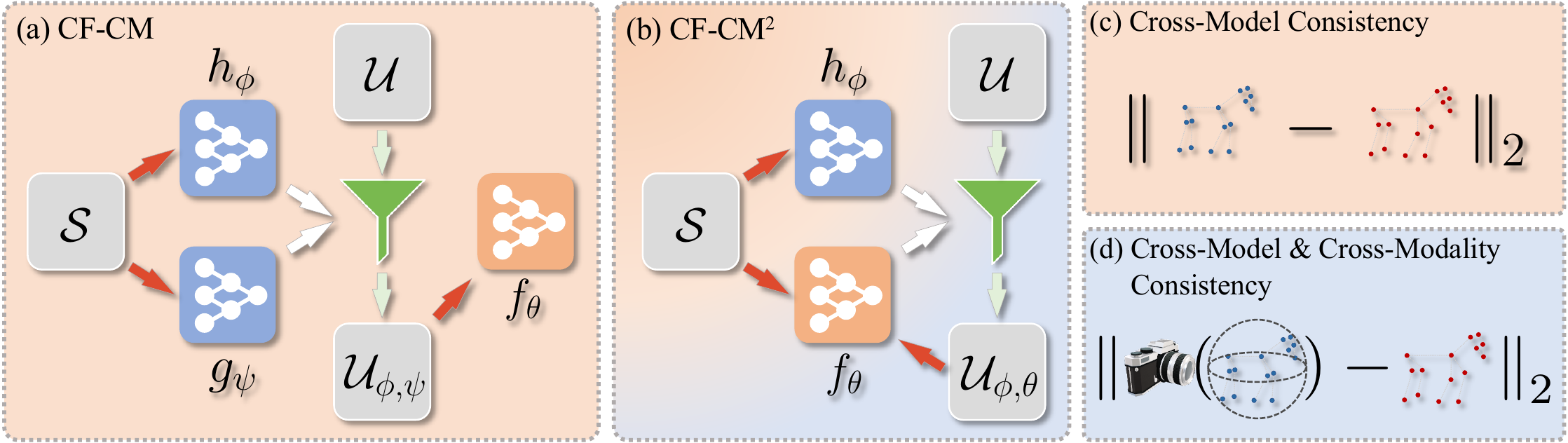}
  \caption{
    Given a small set $\mathcal{S}$ of images labeled with 2D keypoints, we train a 2D keypoint estimation network $h_{\phi}$ and generate keypoint pseudo-labels on web images (set $\mathcal{U}$). We select a subset of $\mathcal{U}$ to train a 3D shape predictor $f_{\theta}$. Two methods for data selection can be seen here: \textbf{(a)} \textbf{CF-CM}: an auxiliary 2D keypoint estimator $g_{\psi}$ generates predictions on $\mathcal{U}$ and images with the smallest discrepancy between the keypoint estimates of $h_{\phi}$ and $g_{\psi}$ are selected (criterion (c)); \textbf{(b)} \textbf{CF-CM$^2$}: $f_{\theta}$ is trained with samples from $\mathcal{S}$ and generates predictions on $\mathcal{U}$. Images with the smallest discrepancy between the keypoint estimates of $h_{\phi}$ and $f_{\theta}$ are selected (criterion (d)) to retrain $f_{\theta}$.
  }
  \label{fig:framework}
\end{figure*}

\section{Introduction}
Predicting the 3D shape of an articulated object from a single image is a challenging task due to its under-constrained nature. Various successful approaches~\cite{kanazawa2018end,kolotouros2019learning} have been developed for inferring the 3D shape of humans. These approaches rely on strong supervision from 3D joint locations acquired using motion capture systems. Similar breakthroughs for other categories of articulated objects, such as animals, remain elusive. This is primarily due to the scarcity of appropriate training data. Some works (such as CMR~\cite{cmr}) learn to predict 3D shapes using only 2D labels for supervision. However, for most object categories even 2D labels are limited or non-existent. We ask: \textit{how can we learn models that predict the 3D shape of articulated objects in-the-wild when limited or no annotated images are available for a given object category?}

In this paper we propose an approach that requires as few as 50-150 images labeled with 2D keypoints. This labeled set can be easily and quickly created for any object category. Our proposed approach is illustrated in Figure~\ref{fig:intro_fig} and summarized as follows: (a)~train a category-specific keypoint estimation network using a small set $\mathcal{S}$ of images labeled with 2D keypoints; (b) generate 2D keypoint pseudo-labels on a large unlabeled set $\mathcal{U}$ consisting of automatically acquired web images; (c) automatically curate $\mathcal{U}$ by creating a subset of images and pseudo-labels $\mathcal{U}'$ according to a selection criterion; (d) train a model for 3D shape prediction with data from both $\mathcal{S}$ and $\mathcal{U}'$.

A key insight is that current 2D keypoint estimators \cite{newell2016stacked,xiao2018simple,sun2019deep} are accurate enough to create robust 2D keypoint detections on unlabeled data, even when trained with a limited number of images. Another insight is that images from $\mathcal{U}$ increase the variability of several factors, such as camera viewpoints, articulations and image backgrounds, that are important for training generalizable models for 3D shape prediction. However, the automatically acquired web images contain a high proportion of low-quality images with wrong or heavy truncated objects. Naively using all web images and pseudo-labels during training leads to degraded performance as can be seen in our experiments in Section~\ref{exps_internet}. While successful pseudo-label (PL) selection techniques~\cite{cascante2021curriculum,chen2021semi,xu2022cross,mu2020learning,li2021synthetic} exist for various tasks, they do not address the challenges in our setting. These works investigate PL selection when the unlabeled images come from curated datasets (\eg CIFAR-10~\cite{cifar10}, Cityscapes~\cite{cordts2016cityscapes}), while in our setting they come from the web. In addition, they eventually use all unlabeled images during training while in our case most of the web images should be discarded. To effectively utilize images from the web we investigate four criteria to automatically create a ``curated" subset that includes images with high-quality pseudo-labels. These contain a confidence-based criterion as well as three consistency-based ones (see Figure~\ref{fig:framework} for two examples).

Through extensive experiments on five different articulated object categories (horse, cow, sheep, giraffe, bear) and three public datasets, we demonstrate that training with the proposed data selection approaches leads in considerably better 3D reconstructions compared to the fully-supervised baseline. Using all pseudo-labels leads to degraded performance. We analyze the performance of the data selection methods used and conclude that consistency-based selection criteria are more effective in our setting. Finally, we conduct experiments with varying number of images in the labeled set $\mathcal{S}$. We show that even with only 50 annotated instances and images from web, we can train models that lead to better 3D reconstructions than the fully-supervised models trained with more labels.
\section{Related Work}

\noindent \textbf{Monocular 3D shape recovery.}
The task of 3D shape recovery from a single image is solved considerably well for the human category, with many approaches achieving impressive results~\cite{kanazawa2018end,kolotouros2019learning,pavlakos2019expressive,kocabas2021pare}. Their success can be attributed to the existence of large datasets with 3D~\cite{ionescu2013human3,mehta2017monocular} and 2D annotations~\cite{andriluka20142d,lin2014microsoft}. However for other articulated object categories, such as animals, datasets with 3D annotations do not exist and even 2D labels are scarce and in some cases not available.

Recent works~\cite{cmr,csm,acsm,li2020self,goel2020shape,kokkinos2021learning} have addressed several aspects of this problem. In CMR~\cite{cmr}, the authors train a system for monocular 3D reconstruction using 2D keypoint and mask annotations. Although their approach works well, it relies on a large number of 2D annotations, such as 6K training samples for 3D reconstruction of birds in CUB~\cite{wah2011caltech}, which limits its direct applicability to other categories. Follow-up works~\cite{li2020self,goel2020shape} attempt to eliminate this reliance by using part segmentations~\cite{li2020self} or mask labels~\cite{goel2020shape}, but their performance is still inferior to CMR~\cite{cmr}, and they require mask or part segmentation annotations during training, which are not always available. Instead, we propose a novel approach that addresses the scarcity of 2D keypoint annotations by effectively utilizing images from the web.

In~\cite{csm,acsm} the authors generate 3D reconstructions in the form of a rigid~\cite{csm} or articulated~\cite{acsm} template, training their models with masks and 2D keypoint labels from Pascal VOC~\cite{everingham2015pascal} and predicted masks (using Mask-RCNN~\cite{he2017mask}) from ImageNet~\cite{deng2009imagenet}. Despite using only a small labeled set and mask pseudo-labels, their approach is not fully automatic as they manually select the predicted masks for their training set. Another line of work~\cite{biggs2020left,ruegg2022barc}, regresses the parameters of a statistical animal model~\cite{zuffi20173d}, but their work only reconstructs the 3D shape of dogs, for which large-scale 2D annotations exist~\cite{biggs2020left}. BARC~\cite{ruegg2022barc} also uses a datasest with 3D annotations during a pre-training stage. Finally, researchers have also utilized geometric supervision in the form of semantic correspondences distilled from images features~\cite{yao2022hi, wu2022magicpony} or temporal signals from monocular videos~\cite{kokkinos2021learning,yang2021lasr,yang2022banmo}.

\noindent \textbf{Learning with keypoint pseudo-labels.}
The use of keypoint pseudo-labels has been widely investigated in various works related to 2D pose estimation~\cite{radosavovic2018data,dong2019teacher,moskvyak2021semi,xie2021empirical,cao2019cross,mu2020learning,li2021synthetic}. In~\cite{radosavovic2018data,xie2021empirical}, the authors investigate learning with keypoint pseudo-labels for human pose estimation, while in~\cite{cao2019cross,mu2020learning,li2021synthetic} the goal is to estimate the 2D pose of animals. In the case of animal pose estimation, PLs are used for domain adaptation from synthetic data~\cite{mu2020learning,li2021synthetic} or from human poses~\cite{cao2019cross}. What is common across these works is that they integrate PLs into the training set (in some cases progressively) and that at the end of the process all samples from the unlabeled set are included. This is not an issue as images in the unlabeled set come from curated datasets. While these approaches can potentially enhance the quality of PLs, a data selection mechanism is still necessary in our case because the unlabeled images come from the web, and most of them should be discarded. To this end, we adapt PL selection criteria form previous works, such as keypoint confidence-based filtering from~\cite{cao2019cross} and multi-transform consistency from~\cite{radosavovic2018data,mu2020learning,li2021synthetic}, to automatically curate the acquired web images.
\section{Approach}

Given a labeled set $\mathcal{S} = \{(I^{(s)}_{i}, x_{i})\}_{1}^{N_{s}}$ comprising of $N_{s}$ images with paired 2D keypoint annotations and an unlabeled set $\mathcal{U} = \{I^{(u)}_{i}\}_{1}^{N_{u}}$ containing $N_{u}$ unlabeled images, our goal is to learn a 3D shape recovery model by effectively utilizing both labeled and unlabeled data. We use a limited annotated set $N_{s}$ and unlabeled images from the web, thus $N_u$ is much larger than $N_s$. We investigate how keypoint pseudo-labeling on unlabeled images can improve existing 3D reconstruction models. CMR~\cite{cmr} and ACSM~\cite{acsm} are chosen as two canoncial examples. Next, we present some relevant background on monocular 3D shape recovery, CMR and ACSM, while in Section~\ref{sec:approach} we present our proposed approach.

\subsection{Preliminaries}
\label{sec:models_3d}
Given an image $I$ of an object, its 3D structure is recovered by predicting a 3D mesh $M$ and camera pose $\pi$ with a model $f_{\theta}$ whose parameters $\theta$ are learned during training.

\noindent \textbf{CMR.}
The shape in CMR~\cite{cmr} is represented as a 3D mesh $M \equiv (V, F)$ with vertices $V \in \mathbb{R}^{|V| \times 3}$ and faces $F$, and is homeomorphic to a sphere. Given an input image $I$, CMR uses ResNet-18~\cite{resnet} to acquire an image embedding, which is then processed by a 2-layer MLP and fed to two independent linear layers predicting vertex displacements $\Delta_{V} \in \mathbb{R}^{|V| \times 3}$ from a template shape $T\in \mathbb{R}^{|V| \times 3}$ and the camera pose. The deformed 3D vertex locations are then computed as $V = T + \Delta_{V}$. A weak-perspective camera model $\pi \equiv (s, \bm{t}, \bm{q})$ is used, where $s \in \mathbb{R_{+}}$ is the scale, $\bm{t} \in \mathbb{R}^{2}$ the translation and $\bm{q} \in \mathbb{R}^{4}$ the rotation of the camera in unit quaternion parameterization. Thus, CMR learns to predict $f_{\theta}(I) \equiv (\Delta_{V}, \pi)$. We refer the reader to~\cite{cmr} for more details.

\noindent \textbf{ACSM.}
The shape in ACSM~\cite{acsm} is also represented as a 3D mesh $M \equiv (V, F)$. The mesh $M$ is a pre-defined template shape for each object category with fixed topology. Given a template shape $T$, its vertices are grouped into $P$ parts. As a result, for each vertex $v$ of $T$ an associated membership $a_{p}^{v} \in [0,1]$ corresponding to each part $p \in \{1, \dots, P\}$ is produced. The articulation $\delta$ of this template $T$ is specified as a rigid transformation w.r.t. each parent part, \textit{i.e.} $\delta \equiv \{(t_p, R_p)\}$, where the torso corresponds to the root part in the hierarchy. Given the articulation parameters $\delta$, a global transformation $\mathcal{T}_{p}(\cdot, \delta)$ for each part can be computed using forward kinematics. Thus, the position of each vertex $v$ of $T$ after articulation can be computed as $\sum_{p} a_{p}^{v} \mathcal{T}_{p} (v, \delta)$. Therefore, to recover the 3D structure of an object with ACSM we need to predict the camera pose $\pi$ and articulation parameters $\delta$. ACSM~\cite{acsm} uses the same image encoder with CMR~\cite{cmr}. The camera pose is parameterized and predicted as in CMR, while independent fully-connected heads regress the articulation parameters $\delta$. Thus, ACSM learns to predict $f_{\theta}(I) \equiv (\delta, \pi)$. For more details, we refer the reader to~\cite{acsm}.

\noindent \textbf{Keypoint and mask supervision.}
\label{sec:kp_3d}
CMR and ACSM are supervised with 2D annotations, by ensuring that the predicted 3D shape matches with the 2D evidence when projected onto the image space using the predicted camera. Let's assume $k$ 2D keypoint locations $x_{i} \in \mathbb{R}^{k \times 2}$ as label for image $I_{i}$. Each one of these $k$ 2D keypoints has a semantic correspondence with a 3D keypoint. The 3D keypoints can be computed from the mesh vertices. In CMR the location of 3D keypoints is regressed from that of the mesh vertices using a learnable matrix $A$, while mesh vertices corresponding to 3D keypoints are pre-determined for ACSM. This leads to $k$ 3D keypoints $\hat{X}_{i} \in \mathbb{R}^{k \times 3}$ that can be projected onto the image using the predicted camera $\hat{\pi}$. In this setting, keypoint supervision is provided using a reprojection loss on the keypoints:
\begin{equation}
    \label{eq:1}
    L_{kp} = ||x_{i} - \hat{\pi}(\hat{X}_{i})||_{2}.
\end{equation}
Similarly, assuming an instance segmentation mask $m_{i}$ is provided as annotation for $I_{i}$, we can render the 3D mesh $M=(V_{i}, F)$ through the predicted camera $\hat{\pi}_{i}$ using a differentiable  renderer~\cite{kato2018neural} $\mathcal{R}(\cdot)$ and provide mask supervision with the following loss:
\begin{equation}
    \label{eq:2}
    L_{mask} = ||m_{i} - \mathcal{R}(V_{i}, F, \hat{\pi}_{i})||_{2}.
\end{equation}
In ACSM, a second network $f'_{\theta'}$ is trained to map each pixel in the object's mask to a point on the surface of the template shape. This facilitates computation of several mask losses. In our experiments with ACSM we only train network $f_{\theta}$ with the keypoint reprojection loss in Eq.~(\ref{eq:1}).

\subsection{3D Shape Recovery Approach}
\label{sec:approach}
Our proposed approach involves the following steps: (1) annotation of 2D keypoints on $N_{s}$ images (when no labels are available for the given category) to create a labeled set $\mathcal{S}$; (2) training a 2D keypoint estimation network with samples from $\mathcal{S}$ and generating 2D keypoint pseudo-labels on an image collection from the web (unlabeled set $\mathcal{U}$); (3) selecting data from $\mathcal{U}$ to be used for training according to a criterion; (4) training a 3D shape predictor with samples from  $\mathcal{S}$ and $\mathcal{U}$. We explain all the steps in detail below.

\noindent \textbf{Data annotation.}
We annotate $N_{S}$ images with 2D keypoints, following standard keypoint definitions used in existing datasets. For instance, we use 16 keypoints for quadruped animals as defined in Pascal~\cite{everingham2015pascal}. The only requirement is to annotate images containing objects with the variety of poses and articulations we wish to model.

\noindent \textbf{Generating keypoint pseudo-labels.}
In pseudo-labeling an initial model is trained with labels from a labeled set $\mathcal{S}$ and is used to produce artificial labels on an unlabeled set $\mathcal{U}$. In this work, instead of tasking the 3D shape predictor $f_{\theta}$ to produce pseudo-labels for itself, we propose to generate 2D keypoint pseudo-labels using the predictions of a 2D keypoint estimation network~\cite{xiao2018simple} $h_{\phi}$. This is a natural option since 2D keypoint estimators are easier to train with limited data and yield more precise detections compared to reprojected keypoints obtained from a 3D mesh.

Most current methods for 2D pose estimation \cite{newell2016stacked,xiao2018simple,sun2019deep} solve the task by estimating $K$ gaussian heatmaps. Each heatmap encodes the probabiblity of a keypoint at a location in the image. The location of each keypoint can be estimated as the location with the maximum value in the corresponding heatmap, while that value can be used as a confidence estimate for that keypoint. We train a 2D keypoint estimation network $h_{\phi}$ with labels from $\mathcal{S}$ and use it to generate keypoint pseudo-labels on $\mathcal{U}$. As such, the unlabeled set is now $\mathcal{U}_{\phi}=\{(I_{i}^{(u)}, \tilde{x}_{i}^{(u)}, \tilde{c}_{i}^{(u)})\}_{1}^{N_u}$, where $\tilde{x}_{i}^{(u)} \in \mathbb{R}^{k \times 2}$ are the estimated locations and $\tilde{c}_{i}^{(u)} \in [0,1]^{k}$ the corresponding confidence estimates for $k$ keypoints of interest in image $I^{(u)}_{i}$.

\noindent \textbf{Learning with keypoint pseudo-labels.}
Given a labeled set $\mathcal{S} = \{(I^{(s)}_{i}, x_{i})\}_{1}^{Ns}$ containing images with keypoint labels $x_{i}$, we train $f_{\theta}$ using the supervised loss:
\begin{equation}
    \label{eq:3}
    L_{kp}^{\mathcal{S}} = \sum_{i=1}^{Ns} ||x_{i} - \hat{\pi}(\hat{X}_{i})||_{2},
\end{equation}
where $\hat{\pi}$ is the predicted camera pose and $\hat{X}_{i}$ the 3D keypoints computed from 3D mesh vertices $V_{i}$. Similarly, given a set $\mathcal{U}_{\phi}=\{(I_{i}^{(u)}, \hat{x}_{i}^{(u)}, \hat{c}_{i}^{(u)})\}_{1}^{N_u}$ comprising of images with keypoint pseudo-labels $(\tilde{x}_{i}, \tilde{c}_{i})$, we train using the following modification of the keypoint reprojection loss:
\begin{equation}
    \label{eq:4}
    L_{kp}^{\mathcal{U}} = \sum_{i=1}^{Nu} ||\tilde{c}_{i}(\tilde{x}_{i} - \hat{\pi}(\hat{X}_{i}))||_{2}.
\end{equation}

We can train using data from both $\mathcal{S}$ and $\mathcal{U}_{\phi}$ with the following objective: $L = L_{kp}^{S} + L_{kp}^{U}$. However, as $Ns \ll N_u$, the training process is dominated by samples from $\mathcal{U}_{\phi}$. When the quality of samples in $\mathcal{U}_{\phi}$ is low, as it is the case with uncurated data from the web, training with all the samples from the unlabeled set leads to degraded performance. To mitigate this issue, we train using a subset of $\mathcal{U}_{\phi}$ containing useful images and high-quality pseudo-labels that are chosen according to a selection criterion.

\subsubsection{Data Selection}
\label{sec:criteria}
Given a number of images $N$ to be used by $\mathcal{U}_{\phi}$, we select them according to one of the following criteria.

\noindent \textbf{Keypoint Confidence.}
This method, referred to as \textit{KP-conf}, selects samples from $\mathcal{U_{\phi}}$ based on the confidence score of the PLs. In particular, it selects $N$ images with with the highest per-sample sum of keypoint confidences $\sum_{k} \tilde{\bm{c}}^{(u)}_{k}$.

\noindent \textbf{Multi-Transform Consistency.}
Intuitively, if one scales or rotates one image the prediction of a good 2D keypoint estimator should change accordingly. In other words, it should be equivariant to the those geometric transformations. We can thus use the consistency to multiple equivariant transformations to select the images with high-quality PLs. We denote this consistency-filtering approach as \textit{CF-MT}.

\noindent \textbf{Cross-Model Consistency.}
Using this consistency filtering approach, referred to as \textit{CF-CM}, we select samples based on the consistency of two 2D keypoint estimators, $h_{\phi}$ and $g_{\psi}$, with different architectures. Our insight is that the two models have different structural biases and can learn different representations from the same image. Thus, we use the discrepancy in their predictions as a proxy for evaluating the quality of the generated pseudo-labels. For each image $I$ from $\mathcal{U}$, we generate keypoint pseudo-labels $\tilde{x}_{\phi}$ and $\tilde{x}_{\psi}$ with $h_{\phi}$ and $g_{\psi}$ respectively, and calculate the discrepancy between the PLs with the following term:
\begin{equation}
    D_{\text{CF-CM}} = ||\tilde{x}_{\phi} - \tilde{x}_{\psi}||_{2}.
\end{equation}
We select $N$ samples with the minimum discrepancy creating the subset $\mathcal{{U}_{\phi,\psi}} \subset \mathcal{U}_{\phi}$.

\noindent \textbf{Cross-Model Cross-Modality Consistency.}
With this consistency filtering criterion, which we dub \textit{CF-CM$^2$}, we select samples based on their discrepancy between the predictions of the 2D keypoint estimator $h_{\phi}$ and the reprojected keypoints from the 3D shape predictor $f_{\theta}$. Given an image $I$ from $\mathcal{U}$, we calculate the locations for 3D keypoints $\tilde{X}_{\theta}$ from the predicted 3D mesh vertices (as explained in Section~\ref{sec:kp_3d}). The corresponding 2D keypoints $\tilde{x}_{\theta}$ are calculated by projecting $\tilde{X}_{\theta}$ onto the image with the predicted camera parameters $\tilde{\pi}_{\theta}$, \textit{i.e.} $\tilde{x}_{\theta} = \tilde{\pi}_{\theta} (\tilde{X}_{\theta})$. As such, we calculate the discrepancy between the keypoints $\tilde{x}_{\phi}$ and $\tilde{x}_{\theta}$ as follows:
\begin{equation}
    D_{\text{CF-CM}^2} = ||\tilde{x}_{\phi} - \tilde{\pi}_{\theta} (\tilde{X}_{\theta})||_{2}.
\end{equation}
Our insight for this consistency check comes from the fact that the two models employ different paradigms for estimating 2D keypoints and have distinct failure modes. 2D keypoint detectors use a bottom-up approach to determine the locations of 2D keypoints. While their predictions are more pixel-accurate than the reprojected keypoints from a 3D mesh, they are not constrained in any way and mainly rely on local cues. Conversely, 3D shape predictors, such as $f_{\theta}$, estimate keypoints as part of a 3D mesh, which constrains their predictions with a specific mesh topology. This makes their predictions more robust.

\begin{figure}[t]
  \centering
  \includegraphics[width=\columnwidth]{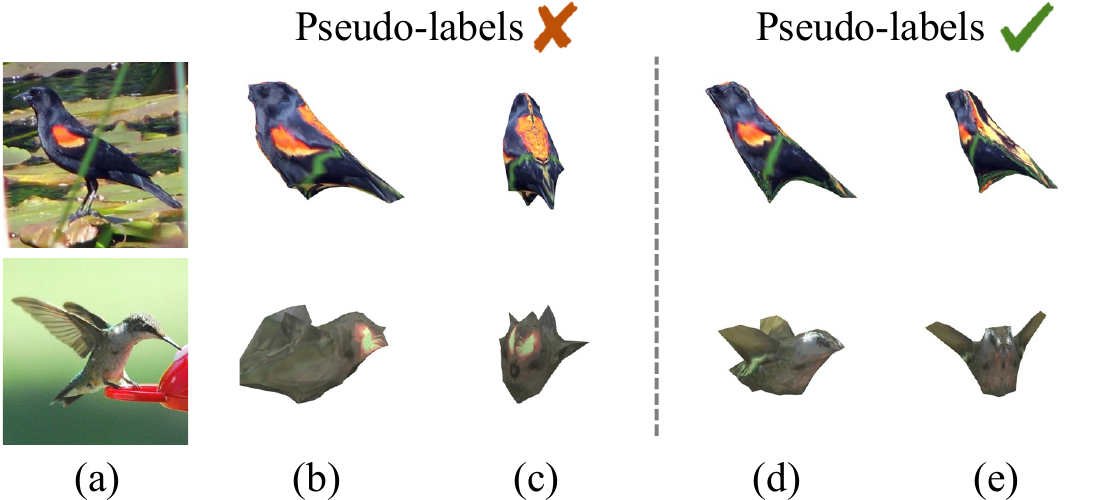}
  \caption{
    {\bf Sample results on CUB}.
    For each sample, we show the predicted shape and texture from the inferred camera view (b, d) and a novel view (c, e) for models trained with and without keypoint pseudo-labels.
  }
  \label{fig:qual_cub}
\end{figure}

\subsection{Additional details}
\label{sec:internet}
Here, we present some additional details of our approach. We include extensive implementation details in the supplementary material.

\noindent \textbf{Data acquisition.}
For each target object category we download images from Flickr using the urls that belong to the public and freely usable YFCC100M~\cite{thomee2016yfcc100m}. However, some images contain wrong objects or objects with heavy truncation (see some examples in Figure~\ref{fig:web_images}). There are also repeated images.

\noindent \textbf{Initial filtering.}
We remove repeated images with dHash~\cite{dhash}. Then, we detect bounding boxes using Mask-RCNN~\cite{he2017mask} with ResNet50-FPN~\cite{resnet,lin2017feature} backbone from \href{https://github.com/facebookresearch/detectron2}{detectron2}. Only detections with confidence above a threshold $\tau$ are retained. We use $\tau=0.95$ for all object categories.

\noindent \textbf{Keypoints from 2D pose estimation network.}
We train the \textit{SimpleBaselines}~\cite{xiao2018simple} pose estimator, with an ImageNet pretrained ResNet-18~\cite{resnet} backbone, with labels from $\mathcal{S}$ and use it to generate keypoint PLs $(\tilde{x}^{(u)}_{\phi}, \tilde{c}^{(u)}_{\phi})$ on $\mathcal{U}$.

\noindent \textbf{Data Selection.}
We curate the web images using a selection criterion. We evaluate the effectiveness of the four selection criteria presented previously. We use each criterion to select the best $N$ data samples from the self-labeled web images. We present experiments with $N \in \{1K, 3K\}$.

\noindent \textbf{Training.}
We train a 3D shape predictor $f_{\theta}$ with keypoint reprojection losses as defined in Eq.~(\ref{eq:3}) \& (\ref{eq:4}).

\section{Experiments}

\label{metrics}
\noindent \textbf{Evaluation metrics.}
While our models are able to reconstruct the 3D shape as a mesh, evaluating their accuracy remains a challenge due to the absence of 3D ground truth data for the current datasets and object categories. Following prior work~\cite{cmr,csm,acsm,kokkinos2021learning} we evaluate our models quantitatively using the standard ``Percentage of Correct Keypoints" (PCK) metric. For $PCK(t)$, the reprojection is ``correct" when the predicted 2D keypoint location is within $t \times max(w, h)$ distance of the ground-truth location, where $w$ and $h$ are the width and the height of the object's bounding box. We summarize the PCK performance at different thresholds by computing the area under the curve (AUC), AUC$_{a_1}^{a_2} =  \int_{a_1}^{a_2} PCK(t) \,dt$. We use $a_1=0.06$ and $a_2=0.1$ and refer to AUC$_{a_1}^{a_2}$ as AUC in our evaluation. Following~\cite{goel2020shape}, we also evaluate the predicted camera viewpoint obtained by our models. We calculate the rotation error $\text{err}_{R} = \arccos\left(\frac{\text{Tr}(\hat{R}^T\tilde{R}) - 1}{2}\right)$ between predicted camera rotation $\hat{R}$ and pseudo ground-truth camera $\tilde{R}$ computed using SfM on keypoint labels. Finally, we offer qualitative results to measure the quality of the predicted 3D shapes.

\begin{table*}[t]
    \centering
    \small

        \begin{tabular}{@{}llcccccc@{}}
            \toprule
            \multicolumn{2}{c}{} & \multicolumn{2}{c}{\textbf{Horse}} & \multicolumn{2}{c}{\textbf{Cow}} & \multicolumn{2}{c}{\textbf{Sheep}}\\
            \cmidrule(lr){3-4}\cmidrule(lr){5-6}\cmidrule(lr){7-8}
           &  &  AUC ($\uparrow$) & err$_{R}$ ($\downarrow$) &  AUC ($\uparrow$) &  err$_{R}$ ($\downarrow$) &  AUC ($\uparrow$) &  err$_{R}$ ($\downarrow$) \\
             
             \midrule
             
            &  ACSM {\scriptsize (Mask)}~\cite{acsm} 
             & 34.9 \reduce{15.9} & 44.3 \reduceb{14.1}
             & 30.7 \reduce{16.9} & 71.0 \reduceb{36.3}
             & 28.2 \reduce{18.6} & 73.1 \reduceb{39.3} 
             \\
           &  ACSM {\scriptsize (KP+Mask)}~\cite{acsm} 
             & 37.4 \reduce{13.4}  & 76.4 \reduceb{46.2}
             & - & -
             & - & -
             \\
           &  ACSM-ours
             & 50.8  & 30.2 
             & 47.6  & 34.7 
             & 46.8  & 33.8 
             \\
           &  ACSM-ours + KP-all
             & 50.2 \reduce{0.6} & 31.9 \reduceb{1.7}
             & 43.9 \reduce{3.9} & 40.1 \reduceb{5.4}
             & 43.5 \reduce{3.5} & 40.7 \reduceb{6.9}
             \\
             
             \midrule
             
             \multirow{4}{*}{\rotatebox[origin=c]{90}{$N=1K$}} & ACSM-ours + KP-conf
             & 53.9 \improve{3.1} & 30.8 \reduceb{0.6}
             & 47.6 & 38.3 \reduceb{3.6}
             & 48.1 \improve{1.3} & 38.6 \reduceb{4.8}
             \\
           &  ACSM-ours + CF-MT
             & 55.1 \improve{4.3} & 30.0 \improveb{0.2}
             & 50.7 \improve{3.1} & 38.9 \reduceb{4.2} 
             & 51.1 \improve{4.3} & 33.3 \improveb{0.5}
             \\
           &  ACSM-ours + CF-CM
             & 54.7 \improve{3.9} & 30.0 \improveb{0.2}
             & 50.9 \improve{3.3} & 38.6 \reduceb{3.9}
             & 50.6 \improve{3.8} & 32.8 \improveb{1.0}
             \\
           &  ACSM-ours + CF-CM$^2$
             & 54.2 \improve{3.4} & 30.9 \reduceb{0.7}
             & 49.4 \improve{1.8} & \textbf{32.7 \improveb{2.0}}
             & 48.6 \improve{1.8} & 32.1 \improveb{1.7}
             \\
             
             \midrule
             
             \multirow{4}{*}{\rotatebox[origin=c]{90}{$N=3K$}} & ACSM-ours + KP-conf
             & 54.3 \improve{3.5}  & 30.0 \improveb{0.2}
             & 50.2 \improve{2.6} & 40.2 \reduceb{5.5}
             & 49.1 \improve{2.3} & 35.7 \reduceb{1.0}
             \\
           & ACSM-ours + CF-MT
             & 55.4 \improve{4.6} & 30.0 \improveb{0.2}
             & 49.2 \improve{1.6} & 39.1 \reduceb{4.4}
             & 51.0 \improve{4.2} & 33.9 \reduceb{0.1}
             \\
           &  ACSM-ours + CF-CM
             & \textbf{55.9 \improve{5.1}} & 29.8 \improveb{0.4}
             & 50.0 \improve{2.4} & 38.8 \reduceb{4.1}
             & 48.4 \improve{1.6} & 34.1 \reduceb{0.3}
             \\ 
           &  ACSM-ours + CF-CM$^2$
             & 55.5 \improve{4.7} & \textbf{28.1 \improveb{2.1}}
             & \textbf{52.6 \improve{5.0}} & 36.2 \reduceb{1.5}
             & \textbf{53.0 \improve{6.2}} & \textbf{31.3 \improveb{2.5}}
             \\
            \bottomrule
        \end{tabular}

    \caption{       
        {\bf Evaluation on Pascal.}
        $N$ is the number of selected images from the web. ACSM-ours + KP-all uses all available web images. Performance change from the fully-supervised baseline ACSM-ours is shown in \textcolor{blue!20!black!50!green}{green}/\textcolor{blue!20!black!30!red}{red}.
    }
	\label{table:pascal}
\end{table*}

\subsection{Simulated Semi-Supervised Setting}
\label{seq:exps_cmr}
First, we investigate the effectiveness of keypoint PLs in a controlled setting. For our experiments we use the CUB dataset~\cite{wah2011caltech} that consists of 6K images with bounding box, mask and 2D keypoint labels for training and testing. We study the impact of training with various levels of supervision on 3D reconstruction performance. We split CUB's training set into a labeled set $\mathcal{S}$ and an unlabeled set $\mathcal{U}$ by randomly choosing samples from the initial training set to be included in $\mathcal{S}$, while placing the rest in $\mathcal{U}$. This is a controlled setting where we expect the performance of the models trained with labels from $\mathcal{S}$ and keypoint pseudo-labels from $\mathcal{U}$ to be lower-bounded by that of models trained using only $\mathcal{S}$ and upper-bounded by the fully-supervised baseline that uses the whole training set. We use CMR~\cite{cmr} for the 3D shape prediction network $f_{\theta}$.

\begin{figure}[t]
  \centering
  \includegraphics[width=\columnwidth]{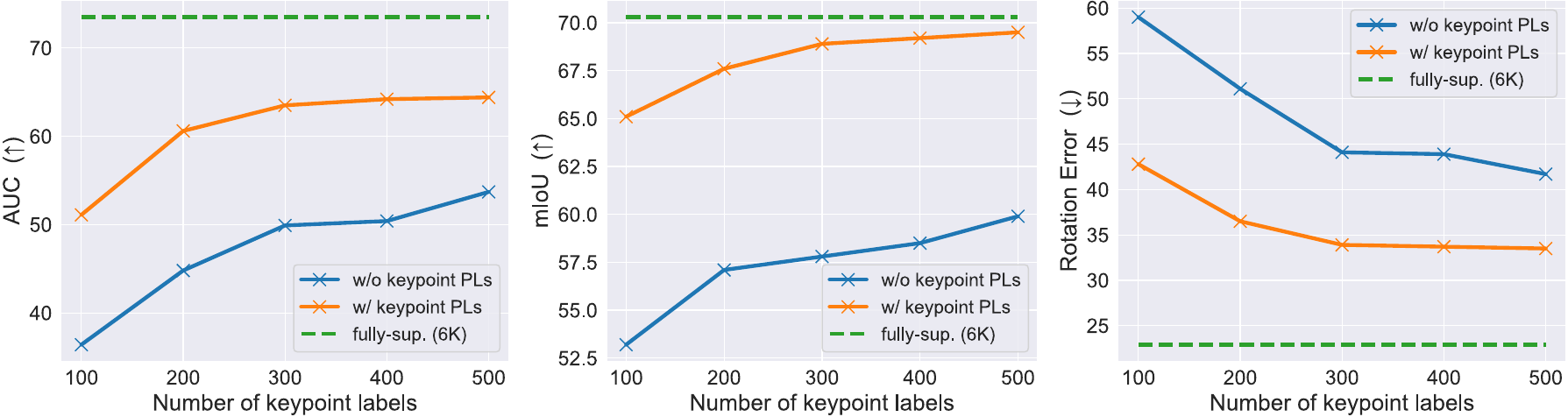}
  \caption{
   \textbf{Evaluation on CUB.} We show the AUC, mIoU and camera rotation error (in degrees) for models trained with different number of labeled and pseudo-labeled instances. Performance across all metrics is significantly improved with keypoint PLs.
  }
  \label{fig:cmr_metrics}
\end{figure}

\noindent \textbf{Results on CUB.}
In Figure~\ref{fig:cmr_metrics}, we compare the performance of models trained with different number of labeled and pseudo-labeled instances. Following CMR, we also report the mIoU metric. We observe that training with 2D keypoint pseudo-labels consistently improves 3D reconstruction performance by a significant margin across all metrics. The improvement is larger when the initial labeled set $\mathcal{S}$ is small (\eg 100 images). This is a very interesting finding. The quality of the generated keypoint pseudo-labels decreases when the 2D keypoint estimator is trained with fewer samples. However, the keypoint pseudo-labels are accurate enough to provide a useful signal for training CMR, largely improving its performance.

In Figure~\ref{fig:qual_cub}, we show sample 3D reconstruction results comparing CMR trained with: i) 300 labeled instances and ii) the same labeled instances and additional keypoint pseudo-labels (for 5,700 images). We show the predicted shape and texture from the predicted camera view and a side view. From Figure~\ref{fig:qual_cub} we can see that the model trained with keypoint pseudo-labels can accurately capture challenging deformations (\eg open wings), while the other model struggles. We offer more qualitative results in the supplementary material.

\subsection{Learning with Web Images}
\label{exps_internet}
We go beyond standard semi-supervised approaches that use unlabeled images from curated datasets by effectively utilizing images from the web for training. We use $N_{S}=150$ images annotated with 2D keypoints and evaluate our proposed approach. 

\noindent \textbf{Datasets.}
We train our models with 2D keypoints from Pascal~\cite{everingham2015pascal} using the same train/test splits as prior work~\cite{csm,acsm}. For categories with less than 150 labeled samples, we augment the labeled training set $\mathcal{S}$ by manually labeling some images from COCO~\cite{lin2014microsoft}. We use Pascal's test split and the Animal Pose dataset~\cite{cao2019cross} for evaluation.

\noindent \textbf{Implementation details.}
We train the \textit{SimpleBaselines}~\cite{xiao2018simple} 2D pose estimator using 150 images with 2D keypoint labels and generate pseudo-labels on web images. For criterion CF-MT, we use scaling and rotation transformations. For criterion CF-CM, we train Stacked HourGlass~\cite{newell2016stacked} as the auxiliary 2D pose estimator. We use ACSM~\cite{acsm} for 3D shape prediction. We found that some template meshes provided in the publicly available ACSM codebase contained incorrect keypoint definitions for certain animals' limbs, with inconsistencies between the right and left sides. After rectifying this issue, we trained ACSM exclusively with keypoint labels and achieved better performance than the one reported in~\cite{acsm}, establishing a stronger baseline. We denote our implementation of ACSM trained with 150 images labeled with 2D keypoints as \textit{ACSM-ours}.

\begin{table*}[t]
    \centering
    \small

        \begin{tabular}{@{}llcccccc@{}}
            \toprule
            \multicolumn{2}{c}{} & \multicolumn{2}{c}{\textbf{Horse}} & \multicolumn{2}{c}{\textbf{Cow}} & \multicolumn{2}{c}{\textbf{Sheep}} \\
            \cmidrule(lr){3-4}\cmidrule(lr){5-6}\cmidrule(lr){7-8}
             &  &  AUC ($\uparrow$) & err$_{R}$ ($\downarrow$) &  AUC ($\uparrow$) &  err$_{R}$ ($\downarrow$) &  AUC ($\uparrow$) &  err$_{R}$ ($\downarrow$) \\

             \midrule
 
           &  ACSM {\scriptsize (Mask)}~\cite{acsm}    
             & 47.4 \reduce{19.3} & 38.0 \reduceb{17.3}
             & 46.2 \reduce{13.4} & 51.8 \reduceb{22.3}
             & 31.4 \reduce{25.8} & 65.6 \reduceb{50.0}
             \\
           &  ACSM {\scriptsize (KP+Mask)}~\cite{acsm} 
             & 51.0 \reduce{15.7} & 59.9 \reduceb{39.2}
             & - & - 
             & - & - 
             \\
           &  ACSM-ours 
             & 66.7 & 20.7
             & 59.6 & 29.5
             & 57.2 & 15.6
             \\
           &  ACSM-ours + KP-all
             & 69.1 \improve{2.4} & 20.9 \reduceb{0.2}
             & 61.2 \improve{1.6} & 27.8 \improveb{1.7}
             & 29.1 \reduce{28.1} & 19.9 \reduceb{4.3}
             \\

             \midrule
             
          \multirow{4}{*}{\rotatebox[origin=c]{90}{$N=1K$}} &  ACSM-ours + KP-conf
             & 72.9 \improve{6.2} & 19.6 \improveb{1.1}
             & 65.0 \improve{5.4} & 31.9 \reduceb{2.4}
             & 58.5 \improve{1.3} & 20.3 \reduceb{4.7}
             \\
           &  ACSM-ours + CF-MT
             & 74.6 \improve{7.9} & 19.9 \improveb{0.8}  
             & 67.0 \improve{7.4} & 26.2 \improveb{3.3}
             & 59.3 \improve{2.1} & \textbf{14.6 \improveb{1.0}}
             \\
          &   ACSM-ours + CF-CM
             & \textbf{75.1 \improve{8.4}} & 19.6 \improveb{1.1}
             & 67.2 \improve{7.6} & 25.9 \improveb{3.6}
             & 59.3 \improve{2.1} & 14.7 \improveb{0.9}
             \\
           &  ACSM-ours + CF-CM$^2$
             & 73.0 \improve{6.3} & 19.7 \improveb{1.0}
             & 67.2 \improve{7.6} & \textbf{23.5 \improveb{6.0}}
             & 58.7 \improve{1.5} & 15.3 \improveb{0.3}
             \\
             
             \midrule
             
          \multirow{4}{*}{\rotatebox[origin=c]{90}{$N=3K$}} &  ACSM-ours + KP-conf
             & 74.3 \improve{7.6} & 19.6 \improveb{1.1}
             & 67.8 \improve{7.8} & 28.0 \improveb{1.5}
             & 57.3 \improve{0.1} & 15.3 \improveb{0.3}
             \\
           &  ACSM-ours + CF-MT
             & 74.4 \improve{7.7} & 19.6 \improveb{1.1}
             & 66.0 \improve{6.4} & 29.7 \reduceb{0.2}
             & 59.6 \improve{2.4} & 15.1 \improveb{0.5}
             \\
           &  ACSM-ours + CF-CM
             & 74.1 \improve{7.4} & 19.6 \improveb{1.1}
             & 66.4 \improve{6.8} & 29.3 \improveb{0.2}
             & 59.4 \improve{2.2} & 15.8 \reduceb{0.2}
             \\ 
           &  ACSM-ours + CF-CM$^2$
             & 73.8 \improve{7.1} & \textbf{19.5 \improveb{1.2}} 
             & \textbf{69.6 \improve{10.0}} & 25.2 \improveb{4.3}
             & \textbf{60.2 \improve{3.0}} & 14.9 \improveb{0.7} 
             \\
            \bottomrule
        \end{tabular}

    \caption{ {\bf Evaluation on Animal Pose.} $N$ is the number of selected images from the web. ACSM-ours + KP-all uses all available web images. Training with selected web images significantly improves the fully-supervised baseline. We show the performance change from ACSM-ours in \textcolor{blue!20!black!50!green}{green}/\textcolor{blue!20!black!30!red}{red}.
    }
	\label{table:animal}
\end{table*}

\begin{figure}[t]
  \centering
  \includegraphics[width=\columnwidth]{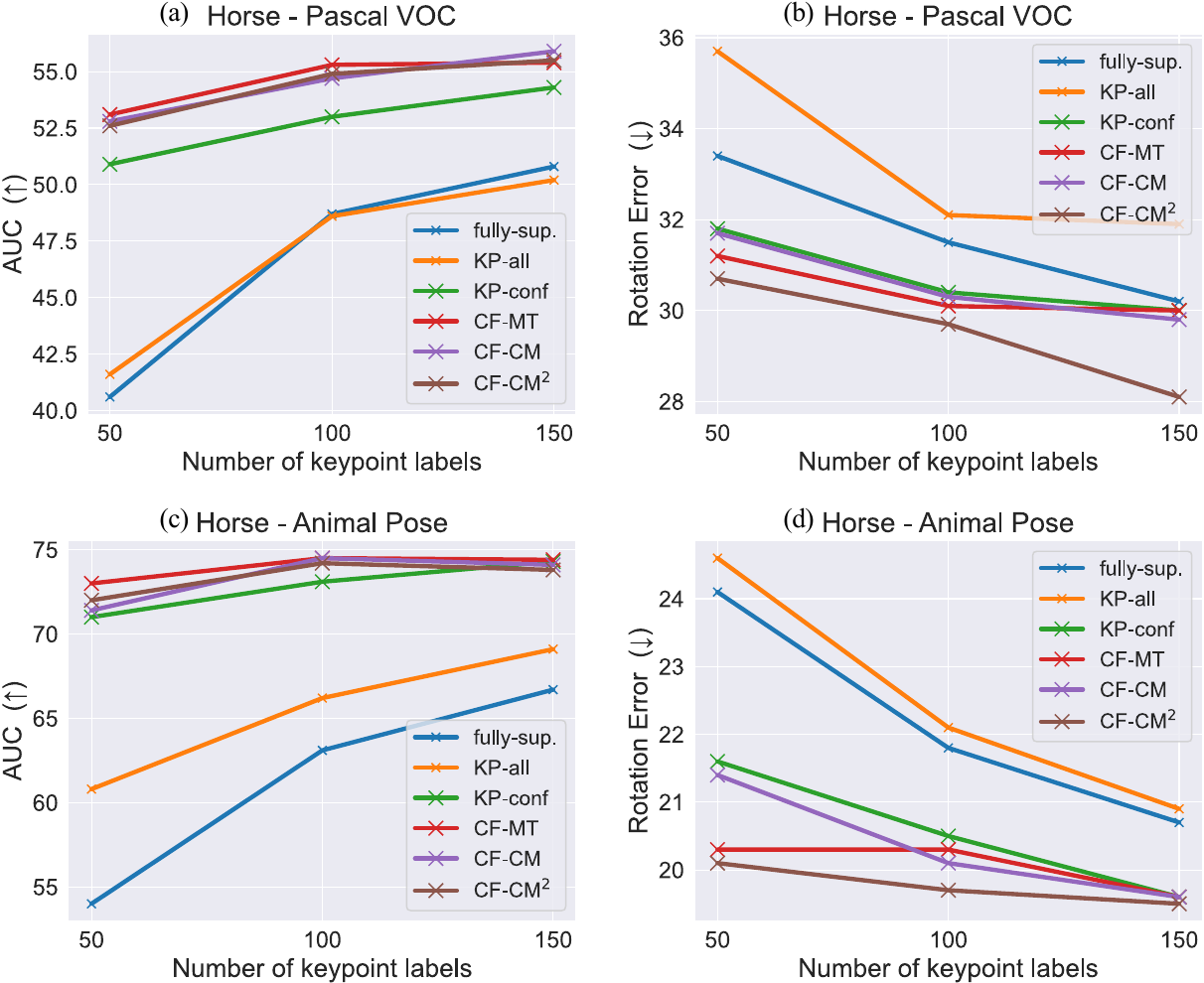}
  \caption{
   \textbf{Evaluation under varying initial supervision.} We show the AUC and camera rotation error (in degrees) on Pascal (top) and Animal Pose (bottom) for the use of our proposed framework under various levels of initial supervision from $\mathcal{S}$. KP-conf, CF-MT, CF-CM, CF-CM$^2$ use PLs from $N=3K$ images.
  }
  \label{fig:horse_abl}
\end{figure}

\begin{table}[t]
	\centering
	\resizebox{\columnwidth}{!}{
 
        \begin{tabular}{@{}llcccc@{}}
            \toprule
            \multicolumn{2}{c}{} & \multicolumn{2}{c}{\textbf{Giraffe}} & \multicolumn{2}{c}{\textbf{Bear}} \\
            \cmidrule(lr){3-4}\cmidrule(lr){5-6}
             &  &  AUC ($\uparrow$) & err$_{R}$ ($\downarrow$) &  AUC ($\uparrow$) &  err$_{R}$ ($\downarrow$) \\
             \midrule
          &   ACSM-ours
             & 79.1 & 34.7
             & 59.2 & \textbf{28.9}
             \\
          &   ACSM-ours + KP-all
             & 75.6 \reduce{3.5} & 39.5 \reduceb{4.8}
             & 58.2 \reduce{1.0} & 35.4 \reduceb{6.5}
             \\
             \midrule
         \multirow{4}{*}{\rotatebox[origin=c]{90}{$N=1K$}} & ACSM-ours + KP-conf
             & 83.3 \improve{4.2} & 31.4 \improveb{3.3}
             & 62.7 \improve{3.5} & 31.0 \reduceb{2.1}
             \\
           &  ACSM-ours + CF-MT
             & 84.4 \improve{5.3} & 32.1 \improveb{2.6}
             & 62.6 \improve{3.6} & 31.9 \reduceb{3.0}
             \\
           &  ACSM-ours + CF-CM
             & 84.1 \improve{5.0} & 32.5 \improveb{2.2}
             & 62.8 \improve{3.6} & 31.6 \reduceb{2.7}
             \\
           &  ACSM-ours + CF-CM$^2$
             & \textbf{86.6 \improve{7.5}} & \textbf{28.8 \improveb{5.9}}
             & 62.2 \improve{3.0} & 30.7 \reduceb{1.8}
             \\
             \midrule
         \multirow{4}{*}{\rotatebox[origin=c]{90}{$N=3K$}} &  ACSM-ours + KP-conf
             & 82.8 \improve{3.7} & 35.6 \reduceb{0.9}
             & 59.9 \improve{0.7} & 35.0 \reduceb{6.1}
            \\
           &  ACSM-ours + CF-MT
             & 83.5 \improve{4.4} & 33.1 \improveb{1.6}
             & 63.1 \improve{3.9} & 31.7 \reduceb{2.8}
             \\
           &  ACSM-ours + CF-CM
             & 83.6 \improve{4.5} & 33.4 \improveb{1.3}
             & \textbf{63.4 \improve{4.2}} & 31.5 \reduceb{2.6}
             \\ 
           &  ACSM-ours + CF-CM$^2$
             & 84.5 \improve{5.4} & 32.8 \improveb{1.9}
             & 62.8 \improve{3.6} & 32.3 \reduceb{3.4}
             \\
            \bottomrule
        \end{tabular}

    }
    \caption{        
        {\bf Evaluation on COCO.}
        $N$ is the number of selected images from the web. ACSM-ours + KP-all uses all available web images. Performance change from the fully-supervised baseline is shown in \textcolor{blue!20!black!50!green}{green}/\textcolor{blue!20!black!30!red}{red}.
    } 
	\label{table:coco}
\end{table}

\noindent \textbf{Baselines.}
We compare the performance of the models trained with selected web images and pseudo-labels with the fully-supervised baseline (\textit{ACSM-ours}) as well as with the baseline that uses all pseudo-labels during training, denoted as \textit{ACSM-ours+KP-all}. To put our results into perspective, we also compare with the models from ACSM~\cite{acsm}. The available models are trained using mask labels from Pascal~\cite{everingham2015pascal} and manually picked mask pseudo-labels from ImageNet~\cite{deng2009imagenet}. For horses, the authors provide a model trained with keypoint labels from Pascal as well.

\begin{figure*}[t]
    \centering
    \includegraphics[width=\textwidth]{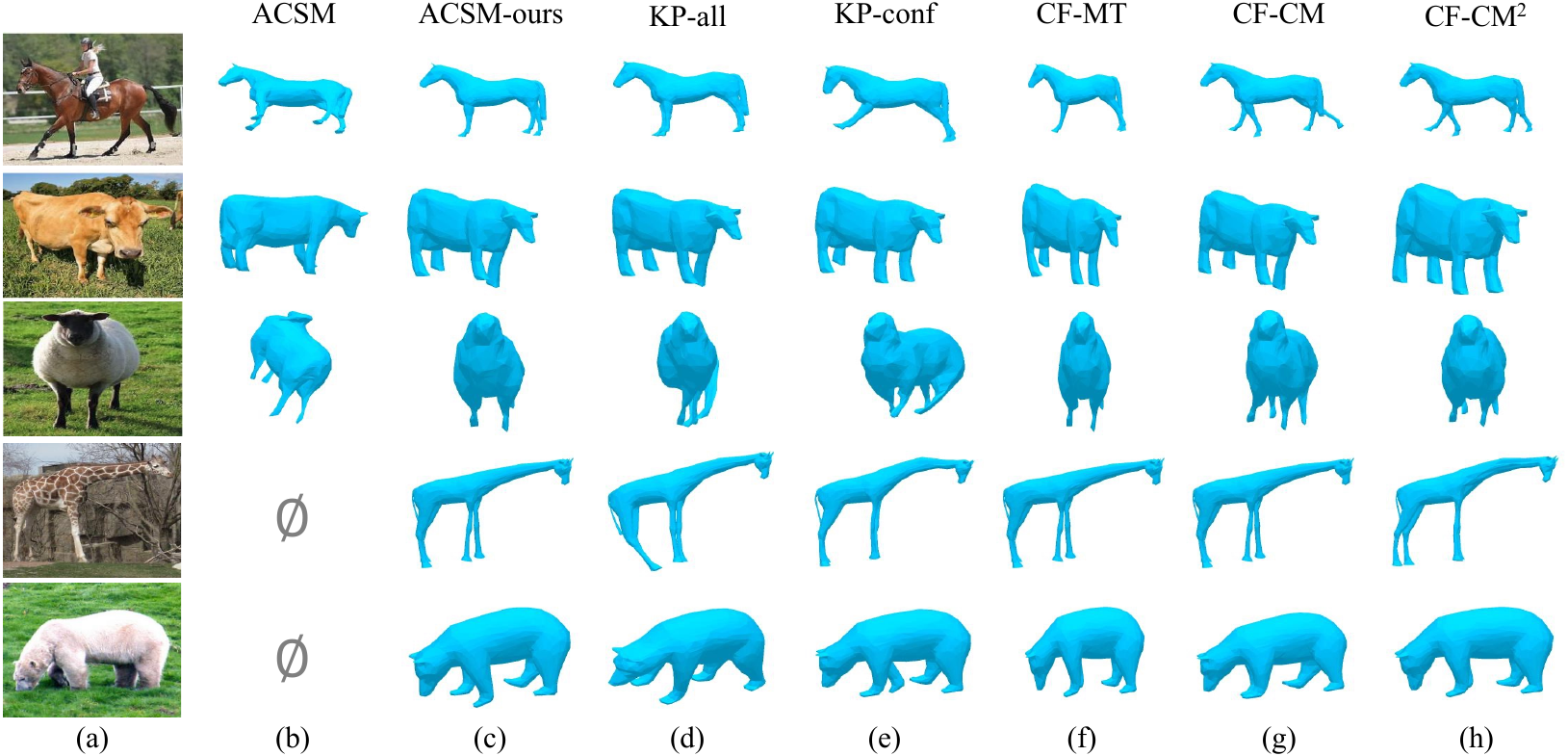}
    \caption{
        {\bf Sample results on 3D shape recovery of quadrupeds.}
        For each input image, we show the predicted 3D shape from the inferred viewpoint. (b) original ACSM (models for giraffes and bears are not available), (c) our ACSM implementation, (d) ACSM-ours trained with all PLs, (e) ACSM-ours + KP-conf, (f) ACSM-ours + CF-MT, (g) ACSM-ours + CF-CM, (h) ACSM-ours + CF-CM$^2$.    
    }
  \label{fig:qual_pa}
\end{figure*}

\noindent \textbf{Results on Pascal.}
First, we evaluate the performance of different methods on Pascal's test set. The results are presented in Table~\ref{table:pascal}. From Table~\ref{table:pascal} we observe that naively including all pseudo-labels during training (ACSM-ours+KP-all) leads in degraded performance. Data selection with KP-conf leads to some improvements, but models trained with pseudo-labels selected from consistency-filtering, \ie CF-MT, CF-CM and CF-CM$^2$, perform consistently better. The improvement over the fully-supervised baseline is quite substantial for those models.  For instance, ACSM-ours+CF-CM$^2$ has a relative improvement (in AUC) of \textbf{10.9\%} over ACSM-ours and \textbf{17.0}\% over ACSM-ours+KP-all, averaged across all categories shown in Table~\ref{table:pascal}.

\noindent \textbf{Results on Animal Pose.}
Next, we evaluate our previously trained models on Animal Pose dataset and present results in Table~\ref{table:animal}. Results are consistent with those in Pascal. Training models with all pseudo-labels results in large performance degradation in some cases (see Sheep in Table~\ref{table:animal}). KP-conf improves over the supervised baseline, but the improvement is higher for CF-MT, CF-CM and CF-CM$^2$. Results from Tables~\ref{table:pascal} \&~\ref{table:animal} suggest that consistency-filtering is more effective than confidence-based filtering in our setting.

\noindent \textbf{Results on COCO.}
In this part, we demonstrate the practicality of our approach in predicting the 3D shape of object categories even in the absence of initial annotations for those categories. We evaluate our approach on giraffes and bears. We annotate 150 images per category from COCO with 2D keypoints and use this set for training. We also annotate some images for evaluation purposes. The annotation process takes less than 1 minute per image. Given a mesh template, we only need to associate the 2D keypoint labels with vertices from the 3D mesh. This process is done in meshlab in $\approx$5 minutes. We train our models with semi-supervised learning as described in Section~\ref{sec:approach}. We compare models that utilize pseudo-labels with the fully-supervised baseline trained with 150 images. The results are presented in Table~\ref{table:coco}. The results are consistent with those on Pascal and Animal Pose. Again, using all pseudo-labels leads in degraded performance. Data selection is essential, and in most cases leads to improved performance compared to the fully-supervised baseline. The performance gains are higher for consistency-based filtering approaches.

\noindent \textbf{Results with different number of labels.}
In Figure~\ref{fig:horse_abl}, we present results on 3D reconstruction of horses with different number $N_{s}$ of images labeled with 2D keypoints. For all experiments, we use $N=3K$ images with pseudo-labels from $\mathcal{U}$ and vary only the number of initial labels $N_{s}$ from $\mathcal{S}$. From Figure~\ref{fig:horse_abl}, we observe that models with keypoint pseudo-labels outperform the fully-supervised baseline even when $N_{s}$ is just 50. More importantly, we notice that models trained with $N_{s}=50$ and keypoint pseudo-labels outperform fully-supervised models with $N_{s}=150$. We also observe, that consistency-based filtering methods are more effective than filtering with keypoint confidence scores. Finally, while all consistency-based filtering approaches lead to similar AUC performance, we notice that CF-CM$^2$ leads to lower camera rotation errors in all cases.

\noindent \textbf{Qualitative Examples.}
In Figure~\ref{fig:qual_pa}, we show sample 3D reconstructions for all models. We observe that ACSM-ours recovers more accurate shapes than the original ACSM version. Training with all PLs results in unnatural articulations in the predicted shapes. Additionally, we observe that some articulations are captured only when training with selected samples through consistency-filtering (\eg horse legs in the first row for CF-CM$^2$). We offer more qualitative results in the supplementary material.
\section{Conclusion}
In summary, we investigated the amount of supervision necessary to train models for 3D shape recovery. We showed that is possible to learn 3D shape predictors with as few as 50-150 images labeled with 2D keypoints. Such annotations are quick and easy to acquire, taking less than 2 hours for each new object category, making our approach highly practical. Additionally, we utilized automatically acquired web images to improve 3D reconstruction performance for several articulated objects, and found that a data selection method was necessary. To this end, we evaluated the performance of four data selection methods and found that training with data selected through consistency-filtering criteria leads to better 3D reconstructions.

\small{
\noindent\textbf{Acknowledgements:} This research has been partially funded by research grants to D. Metaxas through NSF IUCRC CARTA-1747778, 2235405, 2212301, 1951890, 2003874.
}

{\small
\bibliographystyle{ieee_fullname}
\bibliography{ref}
}

\newpage

\setcounter{section}{0}
\renewcommand\thesection{\Alph{section}}
\renewcommand\thesubsection{\thesection.\arabic{subsection}}

\section{Supplementary Material}
In this Supplementary Material we provide additional details that were not included in the main manuscript due to space constraints. In Section B we provide additional implementation details for our experiments. In Section C we offer additional evaluation on hard viewpoints (front \& back view). Finally, we present qualitative results including results on random test samples and failure cases in Section D.

\section{Implementation Details}
\label{sec:implementation}

\noindent \textbf{Pose estimation networks.}
In all of our experiments, we use \textit{SimpleBaselines}~\cite{xiao2018simple} with an ImageNet pretrained ResNet-18~\cite{resnet} backbone as the primary 2D pose estimation network $h_{\phi}$. For criterion CF-CM we train Stacked Hourglass~\cite{newell2016stacked} with 8 stacks for the auxiliary 2D keypoint detector $g_{\psi}$. Both networks use input images of size 256 $\times$ 256 and predict $K$ heatmaps of size 64 $\times$ 64. Random rotations (+/- 30 degrees) and scaling (0.75-1.25) are used as data augmentation. We train both models using Adam~\cite{kingma2015adam} optimizer with learning rate $1 \times 10^{-4}$ for 50K iterations with mini-batches of size 32.

\noindent \textbf{CMR.}
For experiments in Section~\ref{seq:exps_cmr} of the main manuscript, we train CMR~\cite{cmr} with the same hyperparameters as in~\cite{cmr}. We train using Adam~\cite{kingma2015adam} optimizer with learning rate $1 \times 10^{-4}$ for 100K iterations with mini-batches of size 32. In a preprocessing step, CMR  uses SfM on keypoints to initialize the template shape $T$ and acquire a camera estimate for each training instance. We use only the keypoints from $\mathcal{S}$ to initialize $T$. During bundle adjustment, we use the confidence estimate of each keypoint to weight its contribution to the total reprojection error. 

\noindent \textbf{ACSM.}
For experiments in Section~\ref{exps_internet} of the main manuscript, we train ACSM~\cite{acsm}. We only train the network predicting camera poses and articulations from ACSM. We train for 70K iterations with mini-batches of size 12 using Adam~\cite{kingma2015adam} optimizer with learning rate $1 \times 10^{-4}$. We denote training ACSM in this manner as ACSM-ours.


\section{Evaluation on hard viewpoints}
\label{sec:hard_view}

In Table~\ref{table:pascal_hard}, we show results from evaluation on hard viewpoints (front \& back view) in Pascal~\cite{everingham2015pascal}. Those viewpoints are hard for the following reasons: 1) they are not frequent in the training sets (most animals are shown from side views), 2) the instances in those views contain a high degree of self-occlusion. Similar analysis is not applicable to Animal Pose~\cite{cao2019cross} since most instances in that dataset are shown from side views. Results from Table~\ref{table:pascal_hard} suggest that using keypoint pseudo-labels does not only improve 3D reconstruction performance in the mean case. The results are consistent with those in Table 1 \& 2 of the main document, suggesting that consistency-based methods are more effective in our setting.

\begin{table*}[t]
    \centering
        \begin{tabular}{@{}llcccccc@{}}
            \toprule
            \multicolumn{2}{c}{} & \multicolumn{2}{c}{\textbf{Horse}} & \multicolumn{2}{c}{\textbf{Cow}} & \multicolumn{2}{c}{\textbf{Sheep}}\\
            \cmidrule(lr){3-4}\cmidrule(lr){5-6}\cmidrule(lr){7-8}
           &  &  AUC ($\uparrow$) & err$_{R}$ ($\downarrow$) &  AUC ($\uparrow$) &  err$_{R}$ ($\downarrow$) &  AUC ($\uparrow$) &  err$_{R}$ ($\downarrow$) \\
             
             \midrule
             
        & ACSM {\scriptsize (Mask)}~\cite{acsm}
             & 26.7 & 49.1
             & 19.3 & 102.2
             & 14.2 & 94.7
             \\
        &     ACSM {\scriptsize (KP+Mask)}~\cite{acsm}
             & 24.3 & 106.4
             & - & -
             & - & -
             \\
        &     ACSM-ours
             & 45.0 & 35.9
             & 41.7 & 38.5
             & 42.3 & 38.5
             \\
        &     ACSM-ours + KP-all
             & 46.3 & 39.4
             & 39.0 & 42.0
             & 39.9 & 49.8
             \\
             \midrule
        \multirow{4}{*}{\rotatebox[origin=c]{90}{$N=1K$}} & ACSM-ours + KP-conf
             & 47.2 & 37.5
             & 41.6 & 40.6
             & 44.1 & 45.1
             \\
        &     ACSM-ours + CF-MT
             & 47.7 & 37.0
             & 43.2 & 44.4
             & 44.5 & 36.3
             \\
        &     ACSM-ours + CF-CM
             & 47.6 &36.6
             & 43.8 & 42.7
             & \textbf{46.3} & 37.1
             \\
        &     ACSM-ours + CF-CM$^2$
             & 47.5 & 36.6
             & 40.3 & \textbf{36.8}
             & 41.9 & \textbf{32.6}
             \\
             \midrule
        \multirow{4}{*}{\rotatebox[origin=c]{90}{$N=3K$}} & ACSM-ours + KP-conf
             & 46.2 & 36.4
             & 43.6 & 42.0
             & 42.4 & 41.6
             \\
        &     ACSM-ours + CF-MT
             & \textbf{48.3} & 35.3
             & 45.7 & 44.8
             & 44.7 & 40.5
             \\
        &     ACSM-ours + CF-CM
             & 47.8 & 36.1
             & 41.8 & 41.8
             & 44.2 & 36.5
             \\ 
        &     ACSM-ours + CF-CM$^2$
             & 47.7 & \textbf{33.5}
             & \textbf{46.7} & 39.6
             & 45.5 & 33.5
             \\
            \bottomrule
        \end{tabular}

    \caption{       
        {\bf Evaluation on hard viewpoints in Pascal.}
        We report the AUC and camera rotation error err$_{R}$ (in degrees) averaged over images with hard viewpoints (front \& back view). $N$ is the number of selected images from the web.
    }
	\label{table:pascal_hard}
\end{table*}


\section{Additional qualitative results}
\label{sec:qual_supp}

\begin{figure*}[t]
  \centering
  \includegraphics[width=\textwidth]{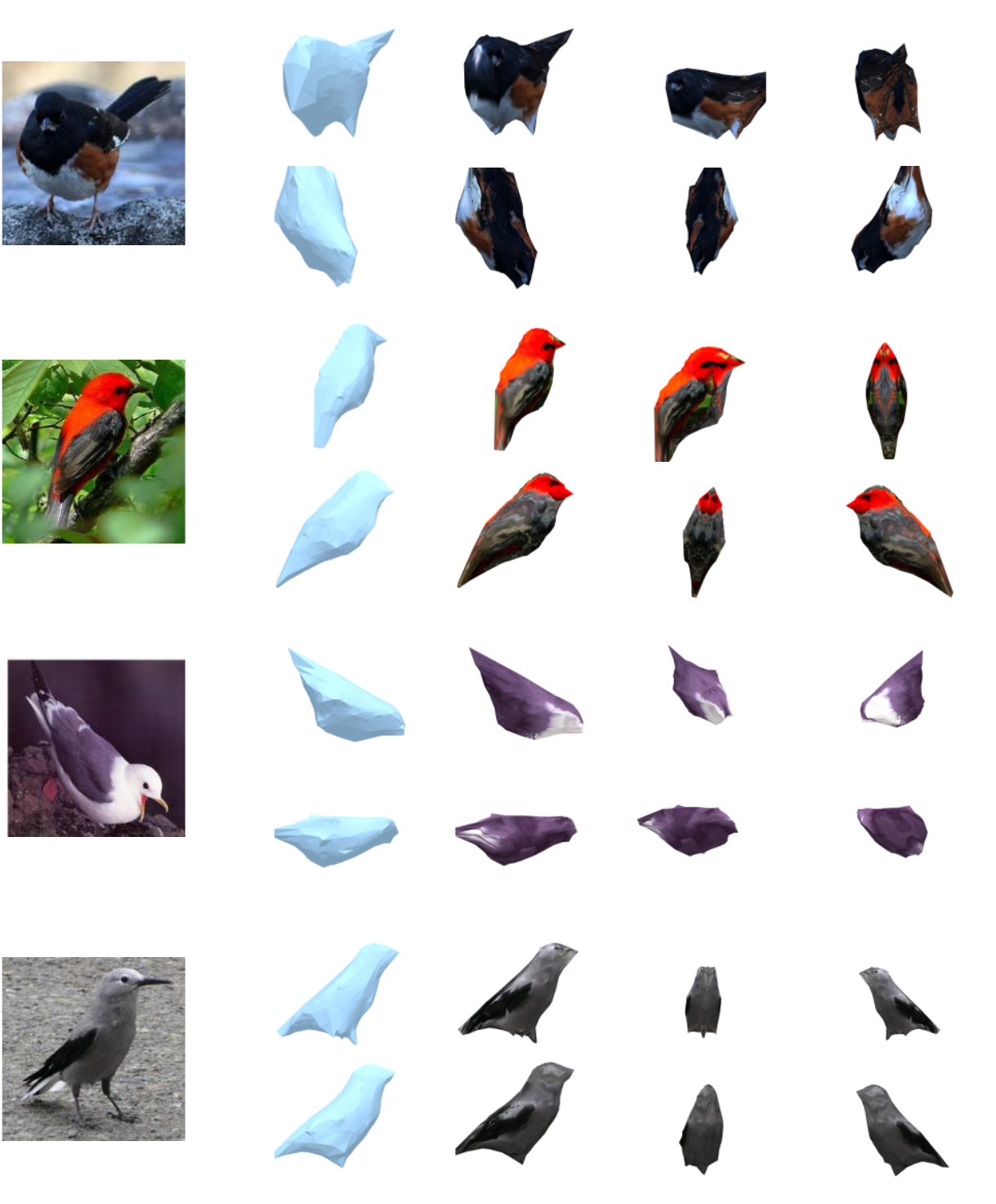}
  \caption{
    {\bf Qualitative results on random samples CUB.}
    For each sample, we compare CMR trained with (first row) and without (second row) keypoint pseudo-labels. The first 2 columns show the predicted shape and texture from the inferred camera viewpoint. The last 2 columns are novel viewpoints of the textured mesh.
  }
  \label{fig:cmr1}
\end{figure*}

\begin{figure*}[t]
  \centering
  \includegraphics[width=\textwidth]{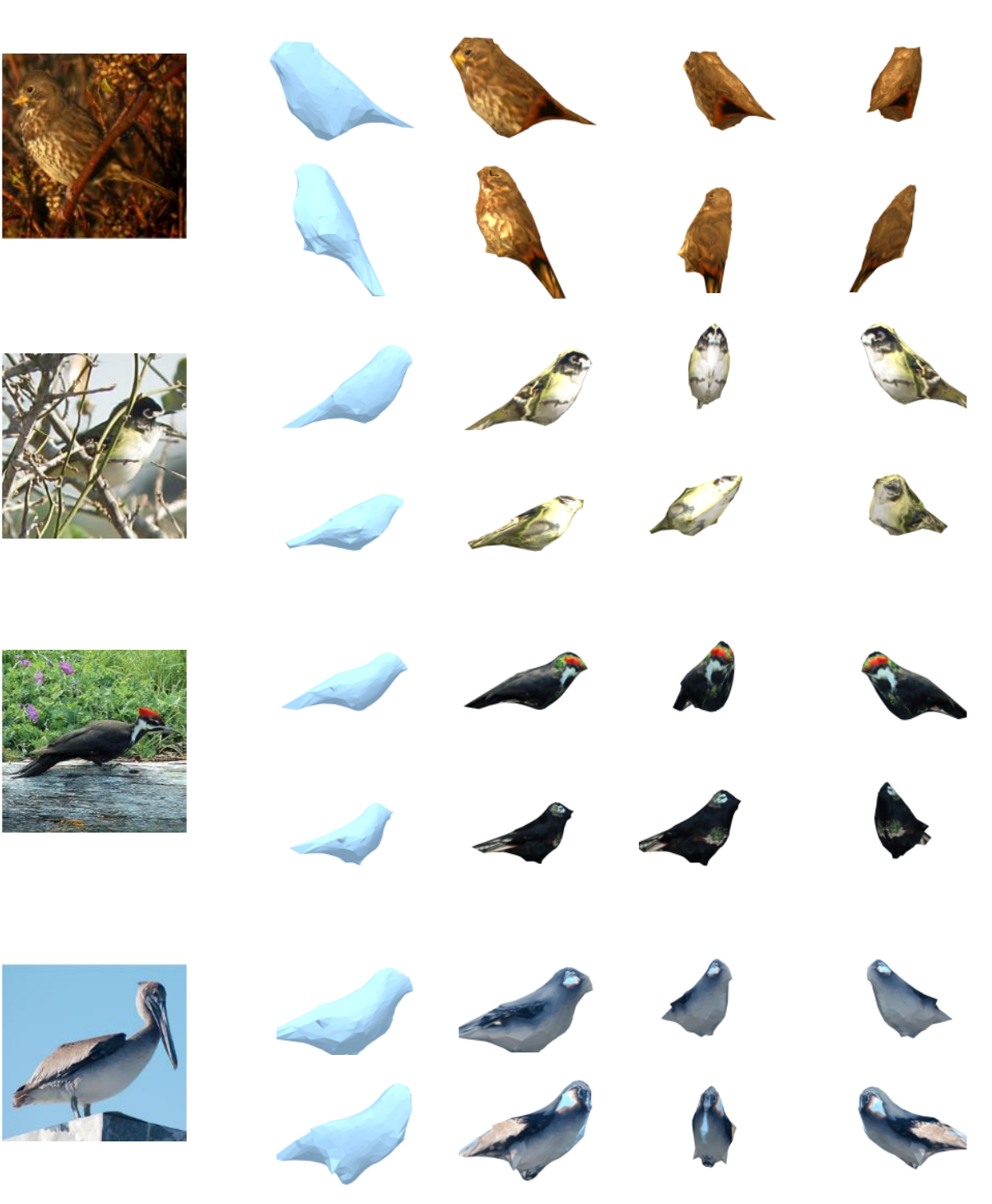}
  \caption{
    {\bf Qualitative results on random samples from CUB.}
    For each sample, we compare CMR trained with (first row) and without (second row) keypoint pseudo-labels. The first 2 columns show the predicted shape and texture from the inferred camera viewpoint. The last 2 columns are novel viewpoints of the textured mesh.
  }
  \label{fig:cmr2}
\end{figure*}

\begin{figure*}[t]
  \centering
  \includegraphics[width=\textwidth]{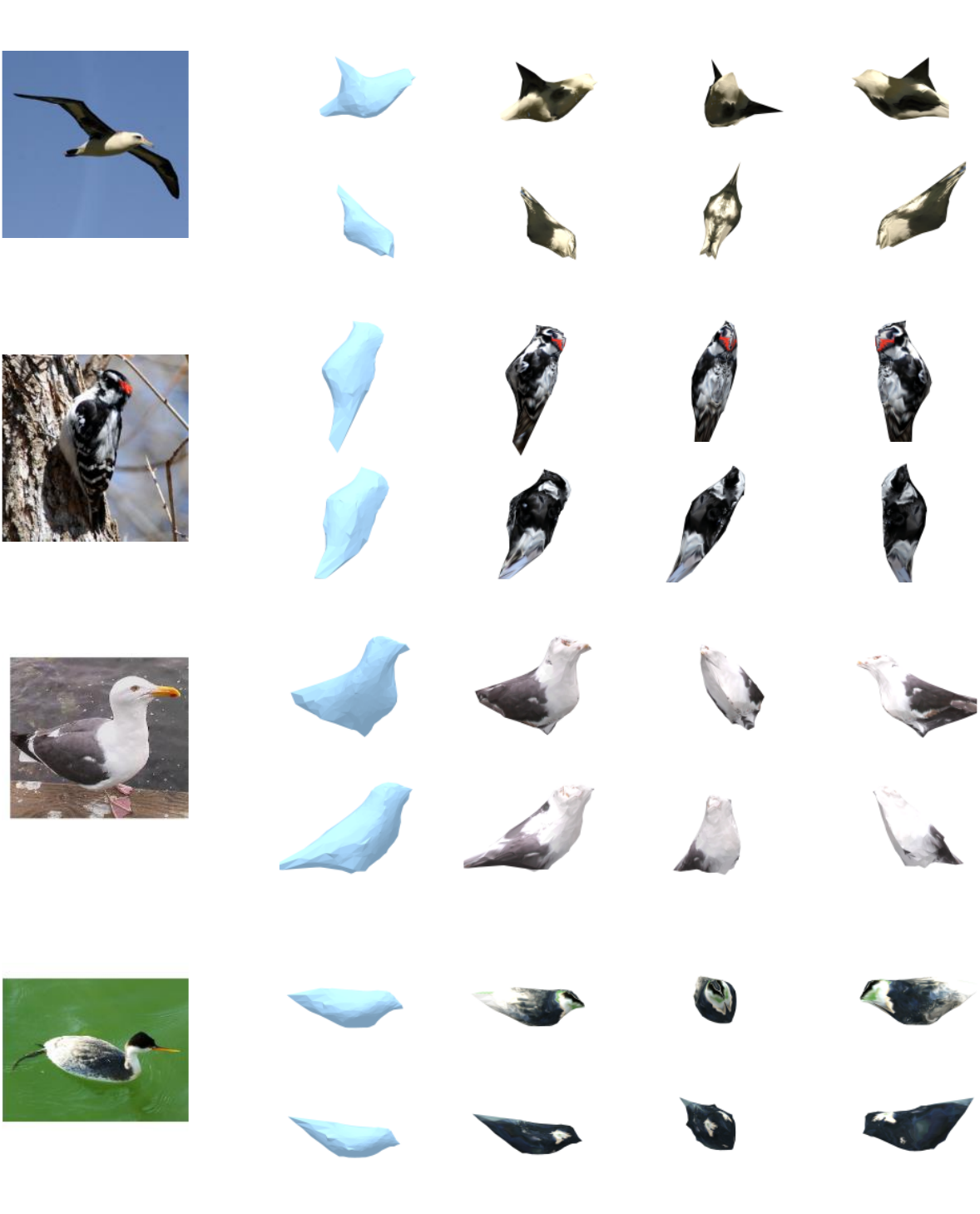}
  \caption{
    {\bf Qualitative results on random samples from CUB.}
    For each sample, we compare CMR trained with (first row) and without (second row) keypoint pseudo-labels. The first 2 columns show the predicted shape and texture from the inferred camera viewpoint. The last 2 columns are novel viewpoints of the textured mesh.
  }
  \label{fig:cmr3}
\end{figure*}

\begin{figure*}[t]
  \centering
  \includegraphics[width=0.75\textwidth]{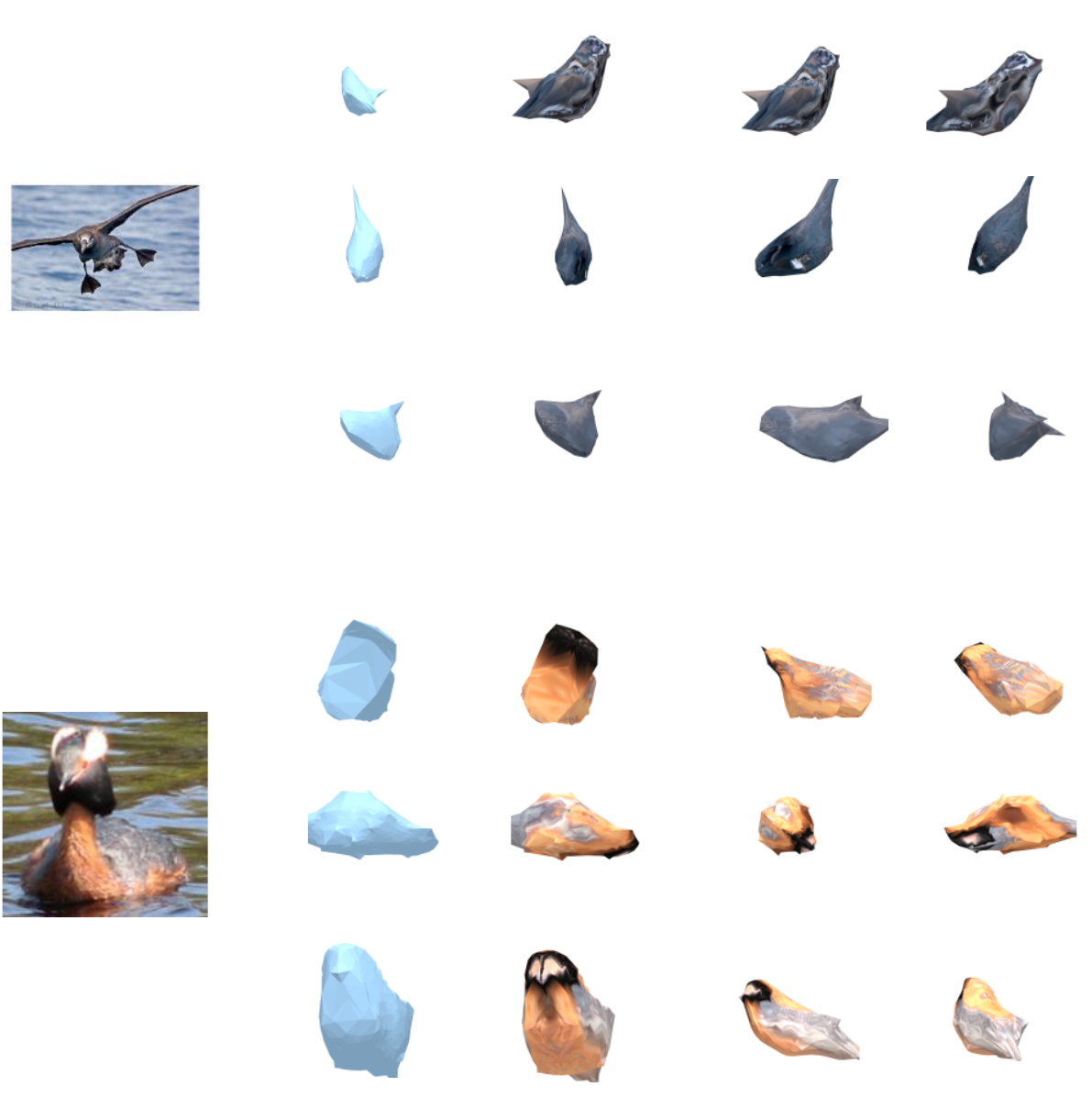}
  \caption{
    {\bf Failure cases for birds.}
    We compare CMR trained with (first row) and without (second row) keypoint pseudo-labels. For a reference, we also show the fully-supervised model trained with 6K mask and keypoints annotations (third row). The first 2 columns show the predicted shape and texture from the inferred camera viewpoint. The last 2 columns are novel viewpoints of the textured mesh. We can see that even the fully-supervised model struggels in those cases.
  }
  \label{fig:cmr_fail}
\end{figure*}

\noindent \textbf{Birds with CMR.}
In Figures~\ref{fig:cmr1}, \ref{fig:cmr2} \&~\ref{fig:cmr3} we compare the predictions of CMR trained with: i) 300 labeled images; ii) the same labeled images and additional keypoint pseudo-labels, using random test samples from CUB. For each input image, the first 2 columns show the predicted shape and texture from the inferred camera viewpoint, while the last 2 columns are novel viewpoints of the textured mesh. We observe that the usage of keypoint pseudo-labels during training significantly improves the 3D reconstruction quality. Training with pseudo-labels enables the model to capture some deformations that the fully-supervised model misses (\eg open wings in the first row of Figure~\ref{fig:cmr3}). These results clearly indicate the merit of using keypoint pseudo-labels with CMR. Finally, we visualize some failure cases of the proposed method in Figure~\ref{fig:cmr_fail}. In those cases even the CMR model supervised with all the 6K images from the training set struggles.

\begin{figure*}[t]
  \centering
  \includegraphics[width=\textwidth]{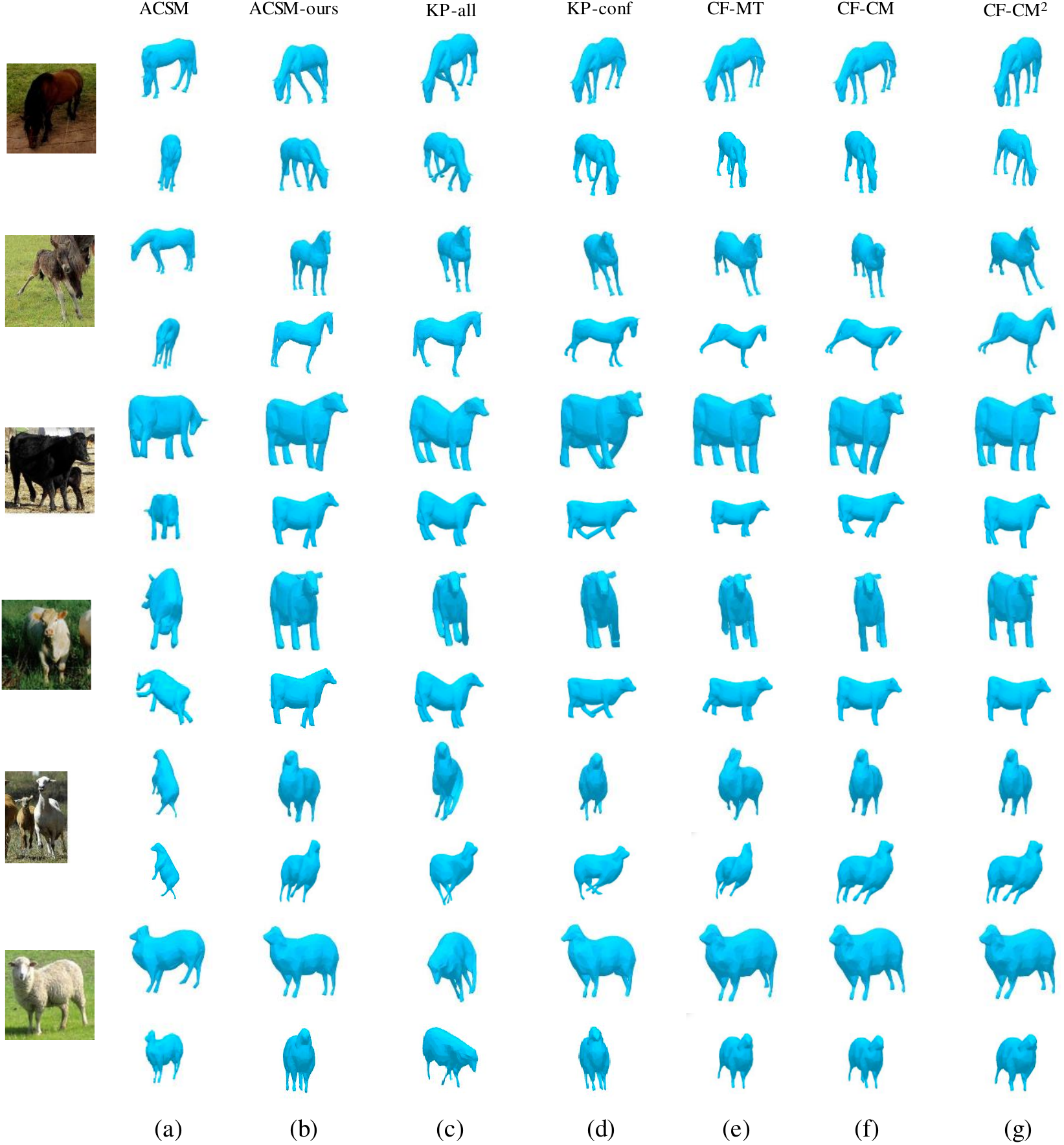}
  \caption{
    {\bf Qualitative results for quadrupeds.}
    Qualitative comparisons between all methods using images from Pascal's test set. For each input image, we show the articulated shape from the inferred camera viewpoint (top row) and a side view (bottom row).
  }
  \label{fig:acsm1}
\end{figure*}

\begin{figure*}[t]
  \centering
  \includegraphics[width=\textwidth]{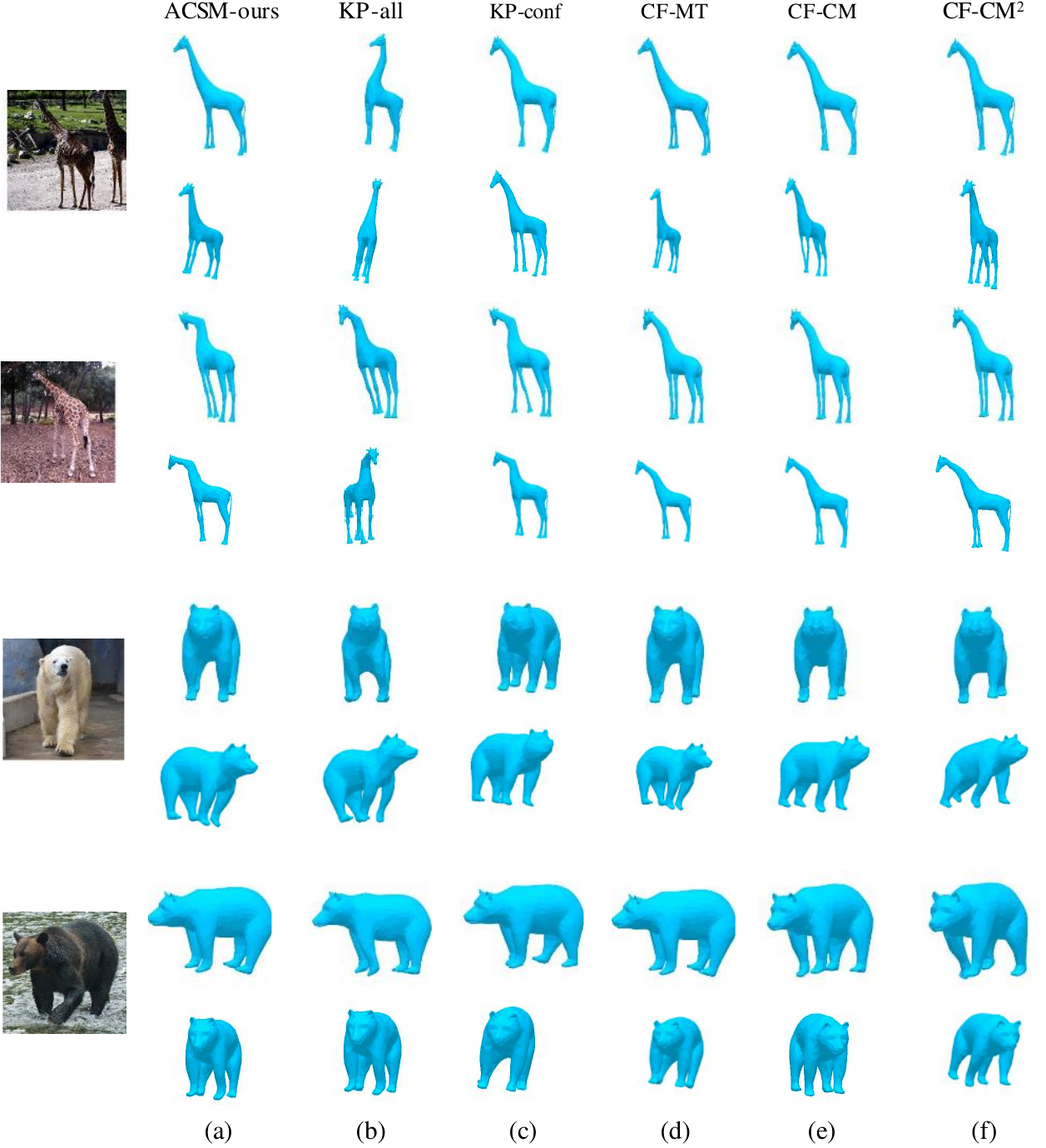}
  \caption{
    {\bf Qualitative results for quadrupeds.}
    Qualitative comparisons between all methods using images from COCO. For each input image, we show the articulated shape from the inferred camera viewpoint (top row) and a side view (bottom row).
  }
  \label{fig:acsm2}
\end{figure*}

\noindent\textbf{Quadrupeds with ACSM.}
In Figures~\ref{fig:acsm1} \& \ref{fig:acsm2} we show qualitative comparisons between all the methods used for predicting the 3D shape of quadrupeds. For each input image, we show the predicted 3D mesh from the inferred camera view (first row) and a novel view (second row). We show results with $N=3K$ samples from $\mathcal{U}$ for KP-conf, CF-MT, CF-CM and CF-CM$^2$. From Figures~\ref{fig:acsm1} \& \ref{fig:acsm2} we observe that the quality of the predicted 3D shapes is consistent with the quantitative evaluation conducted in the main maniscript (Tables 1, 2, 3). First, we observe that ACSM-ours (trained only with 150 images with keypoint-labels) achieves more accurate reconstructions than ACSM for all categories. A failure mode of ACSM is erroneous camera pose prediction. With ACSM-ours camera poses are improved, but some articulations are not well captured by the model. Using supervision from all web images (KP-all) increases errors in camera poses and results in unnatural articulations (see the prediction for sheep in Figure~\ref{fig:acsm1}). KP-conf improves the quality of the predicted shapes compared to KP-all, but still results in unnatural articulations in some cases (see the second giraffe's neck at Figure~\ref{fig:acsm2}). Finally, consistency-based filtering can lead in more accurate camera pose and articulation prediction than other alternatives. For instance, the articulation of the last bear in Figure~\ref{fig:acsm2} is only captured by CF-CM and CF-CM$^2$.

\begin{figure*}[t]
  \centering
  \includegraphics[width=\textwidth]{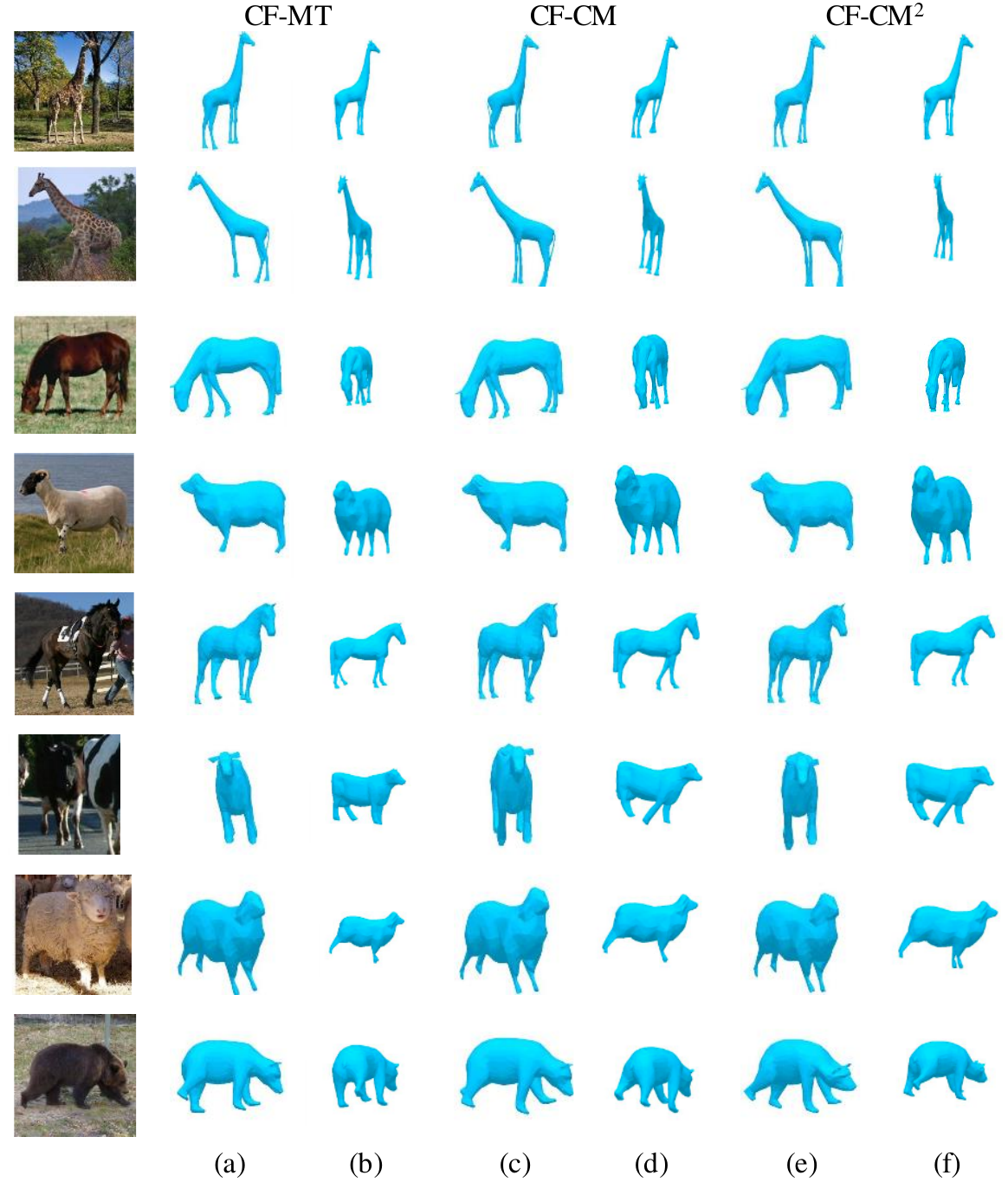}
  \caption{
    {\bf Qualitative results on random samples for quadrupeds.}
    We visualize the recovered 3D shape from the predicted (a, c, e) and a novel view (b, d, f) for models trained with data selected from consistency-based criteria.
  }
  \label{fig:acsm3}
\end{figure*}

\begin{figure*}[t]
  \centering
  \includegraphics[width=\textwidth]{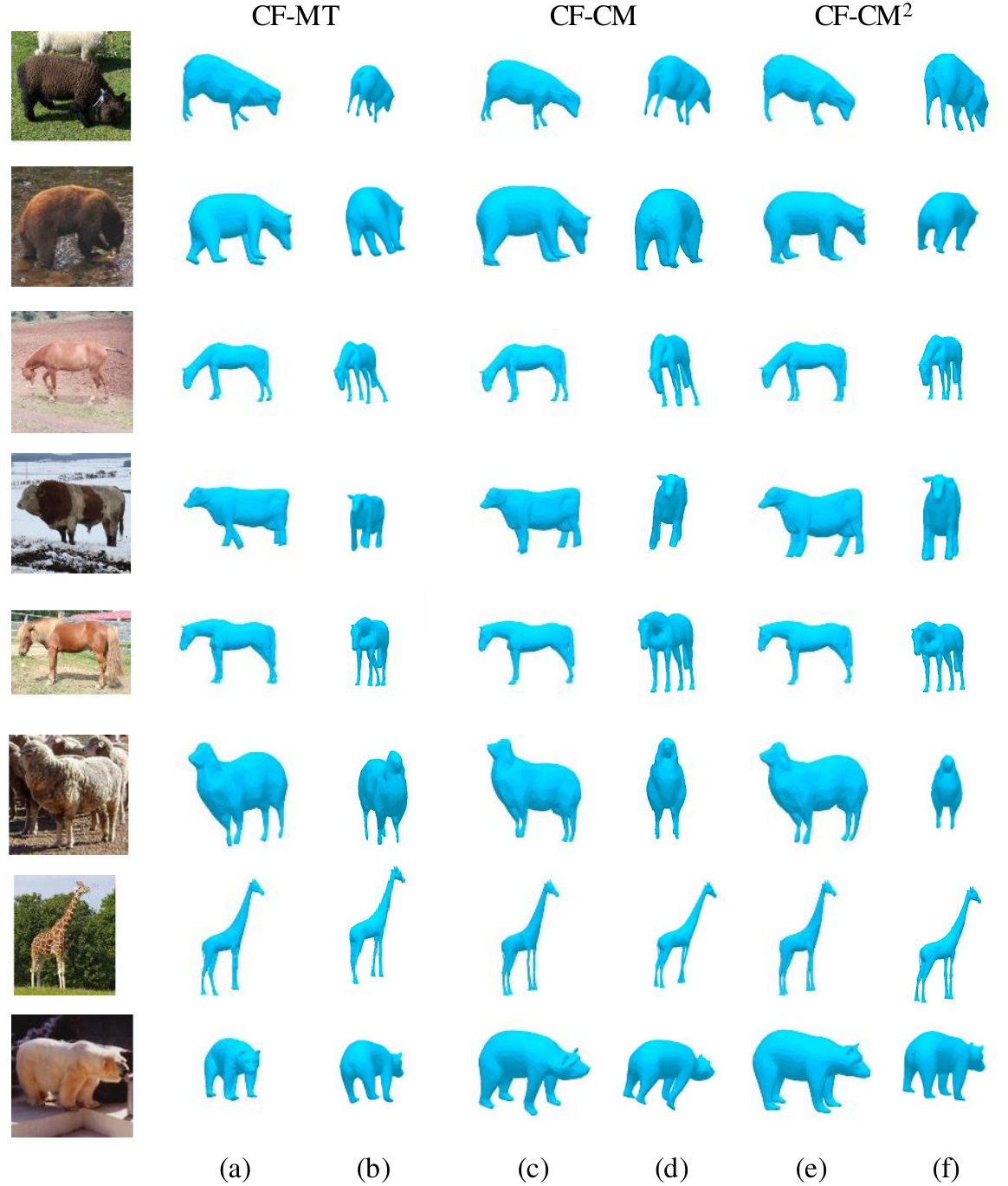}
  \caption{
    {\bf  Qualitative results on random samples for quadrupeds.}
    We visualize the recovered 3D shape from the predicted (a, c, e) and a novel view (b, d, f) for models trained with data selected from consistency-based criteria.
  }
  \label{fig:acsm4}
\end{figure*}

In Figures~\ref{fig:acsm3} \& \ref{fig:acsm4} we visualize the recovered 3D shape for models trained with data selected using consistency-filtering criteria. For each input image, we show the recovered 3D shape from the predicted and a novel view. As also captured by the quantitative evaluation in our main paper, we observe that the quality of the recovered 3D shape is in some cases higher for CF-CM$^2$ compared to other alternatives.

\begin{figure*}[h]
  \centering
  \includegraphics[width=\textwidth]{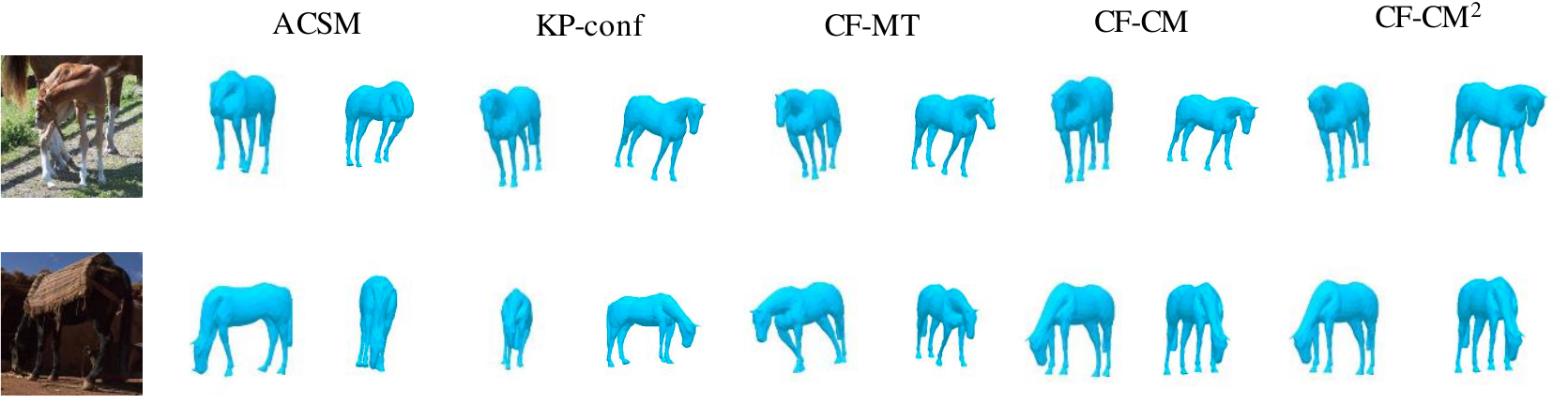}
  \caption{
    {\bf Failure cases for quadrupeds.}
    For each input image, we show the predictions from the inferred (left column) and a novel (right column) view. Common failure modes include errors in articulations (top row) that the ACSM model is not possible to capture by design, and erroneous camera pose prediction for hard viewpoints.
  }
  \label{fig:acsm_fail}
\end{figure*}

In Figure~\ref{fig:acsm_fail}, we visualize some failure cases. We also include the prediction of the original ACSM model. Common failure modes include the inability to capture certain articulations (top row). These articulations are impossible to be captured by ACSM since it models articulations as rigid transformations of some pre-defined parts. Another failure mode is erroneous camera pose prediction for hard viewpoints (second row). ACSM fails worse in those cases (see unnatural head prediction in top row of Figure~\ref{fig:acsm_fail}).

\section{Sample web images from Flickr}
\begin{figure*}[t]
  \centering
  \includegraphics[width=0.95\textwidth]{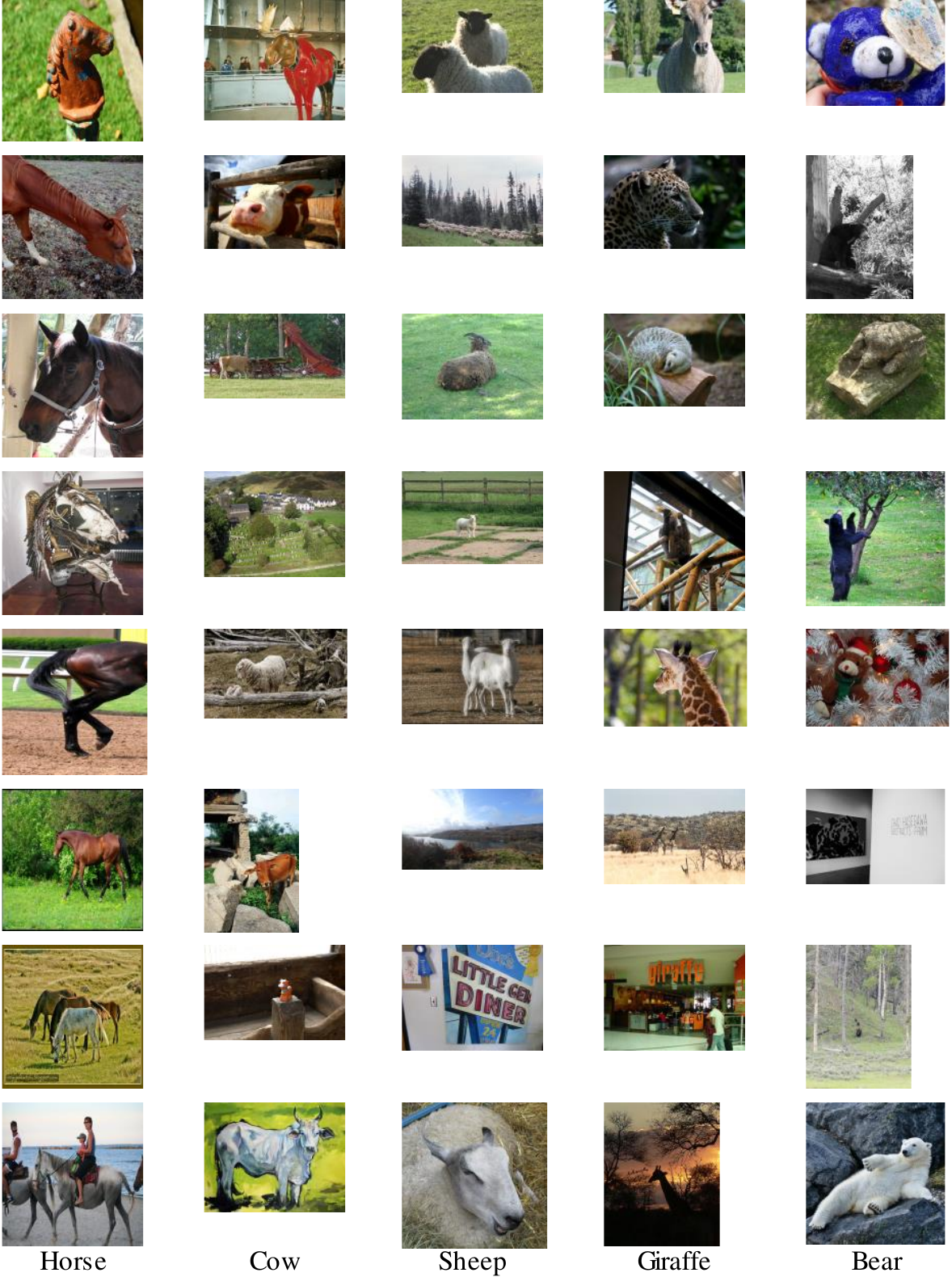}
  \caption{
    We randomly sample 8 images per object category from the unlabeled web images to stress the necessity for an effective data selection mechanism in our setting.
  }
  \label{fig:web_images}
\end{figure*}

In Figure~\ref{fig:web_images} we show 8 random samples per object category from the unlabeled web images. Most images are not suitable from training 3D shape prediction models and should be filtered out.

\end{document}